\documentclass[onecolumn, 10pt]{article}

\newcommand{\be}{\begin{equation}}
\newcommand{\ee}{\end{equation}}
\newcommand{\beq}{\begin{equation}}
\newcommand{\eeq}{\end{equation}}
\newcommand{\bed}{\begin{displaymath}}
\newcommand{\eed}{\end{displaymath}}
\newcommand{\beqa}{\begin{eqnarray}}
\newcommand{\eeqa}{\end{eqnarray}}
\newcommand{\beqann}{\begin{eqnarray*}}
\newcommand{\eeqann}{\end{eqnarray*}}
\newcommand{\bseq}{\begin{subequations}}
\newcommand{\eseq}{\end{subequations}}

\newcommand{\ba}{\begin{array}}
\newcommand{\ea}{\end{array}}

\newcommand{\negr}[1]{{\bf {#1}}}

\usepackage[dvips]{graphicx}
\usepackage{psfrag}
\usepackage{epsfig}
\usepackage{array}
\usepackage{amssymb}
\usepackage{subeqn}
\usepackage{subfigure}

\title{The Design of a Novel Prismatic Drive \goodbreak for a Three-DOF Parallel-Kinematics Machine}
\author{D. Chablat$^1$, J. Angeles$^2$ \\
      $^1$Institut de Recherche en Communications et Cybern\'etique de
      Nantes
      \thanks{IRCCyN: UMR n$^\circ$ 6597 CNRS, \'Ecole Centrale de Nantes,
                        Universit\'e de Nantes, \'Ecole des Mines de
                        Nantes}\\
      UMR CNRS n$^\circ$ 6597, 1 rue de la No\"e, 44321 Nantes, France \\
      $^2$Department of Mechanical Engineering \& \\
      Centre for Intelligent Machines, McGill University \\
      817 Sherbrooke Street West, Montreal, Canada H3A 2K6 \\
      Damien.Chablat@irccyn.ec-nantes.fr $\quad$ angeles@cim.mcgill.ca
    }
\begin{document}
\maketitle
\begin{abstract}
The design of a novel prismatic drive is reported in this paper. This transmission is based on {\it Slide-o-Cam}, a cam mechanism with multiple rollers mounted on a  common translating follower. The design of Slide-o-Cam was reported elsewhere. This drive thus provides pure-rolling motion, thereby reducing the friction of rack-and-pinions and linear drives. Such properties can be used to design new transmissions for parallel-kinematics machines. In this paper, this transmission is intended to replace the ball-screws in Orthoglide, a three-dof parallel robot intended for machining applications.
\end{abstract}
\section{Introduction}
In robotics and mechatronics applications, whereby motion is controlled using a piece of software, the conversion of motion from rotational to translational is usually done by either {\em ball-screws} or {\em linear actuators}. Of these alternatives, ball-screws are gaining popularity, one of their drawbacks being the high number of moving parts that they comprise, for their functioning relies on a number of balls rolling on grooves machined on a shaft; one more drawback of ball-screws is their low load-carrying capacity, stemming from the punctual form of contact by means of which loads are transmitted. Linear bearings solve these drawbacks to some extent, for they can be fabricated with roller-bearings; however these devices rely on a form of direct-drive motor, which makes them expensive to produce and to maintain.

A novel transmission, called {\it Slide-o-Cam}, was introduced in
\cite{Gonzalez-Palacios:2000} as depicted in Fig.~\ref{fig001}, to
transform a rotation into a translation. Slide-o-Cam is composed of 
four major elements: ($i$) the frame, ($ii$) the
cam, ($iii$) the follower, and ($iv$) the rollers. The input axis
on which the cam is mounted, the camshaft, is driven at a constant
angular velocity by an actuator under computer-control. Power is
transmitted to the output, the translating follower, which is the
roller-carrying slider, by means of pure-rolling contact between
cam and roller. The roller comprises two components, the pin and
the bearing. The bearing is mounted at one end of the pin, while
the other end is press-fit into the roller-carrying slider.
Contact between cam and roller thus takes place at the outer
surface of the bearing. The mechanism uses two conjugate
cam-follower pairs, which alternately take over the motion
transmission to ensure a positive action; the rollers are thus
driven by the cams throughout a complete cycle. The main advantage
of cam-follower mechanisms over alternative
transmissions to transform rotation into translation is that
contact through a roller reduces friction, contact stress and
wear.

This transmission, once fully optimized, will replace the three
ball-screws used by the Orthoglide prototype \cite{Chablat:2003}.
Orthoglide features three prismatic joints mounted orthogonaly,
three identical legs and a mobile platform, which moves in
Cartesian space with fixed orientation, as shown in
Fig.~\ref{fig002}. The three motors are SANYO DENKI (ref.
P30B08075D) with a constant torque of 1.2~Nm from 0 to 3000~rpm.
This property enables the mechanism to move throughout the
workspace a 4~kg load with an acceleration of 17~ms$^{-2}$ and a
velocity of 1.3~ms$^{-1}$. On the ball-screws, the pitch is 50~mm
per cam turn. The minimum radius of the camshaft is 8.5~mm. A new
arrangement of camshaft, rollers and follower is proposed to reduce the
inertial load when more than two cams are used.

 {\begin{figure}[htb]
 \begin{minipage}[b]{8cm}
 \begin{center}
   \psfrag{Roller}{Roller}
   \psfrag{Follower}{Follower}
   \psfrag{Conjugate cams}{Conjugate cams}
   \centerline{\epsfig{file = 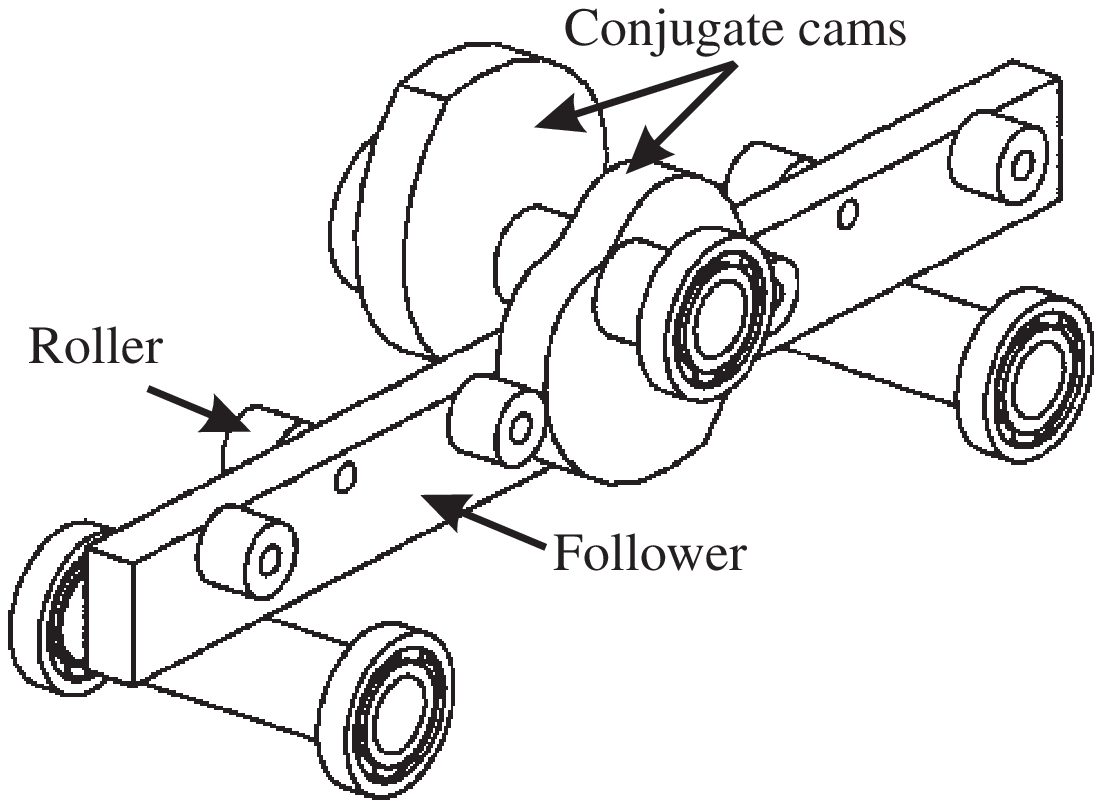,scale = 0.5}}
   \caption{Layout of \goodbreak Slide-o-Cam}
   \label{fig001}
 \end{center}
 \end{minipage}
 \begin{minipage}[b]{8cm}
 \begin{center}
    \centerline{\epsfig{file = 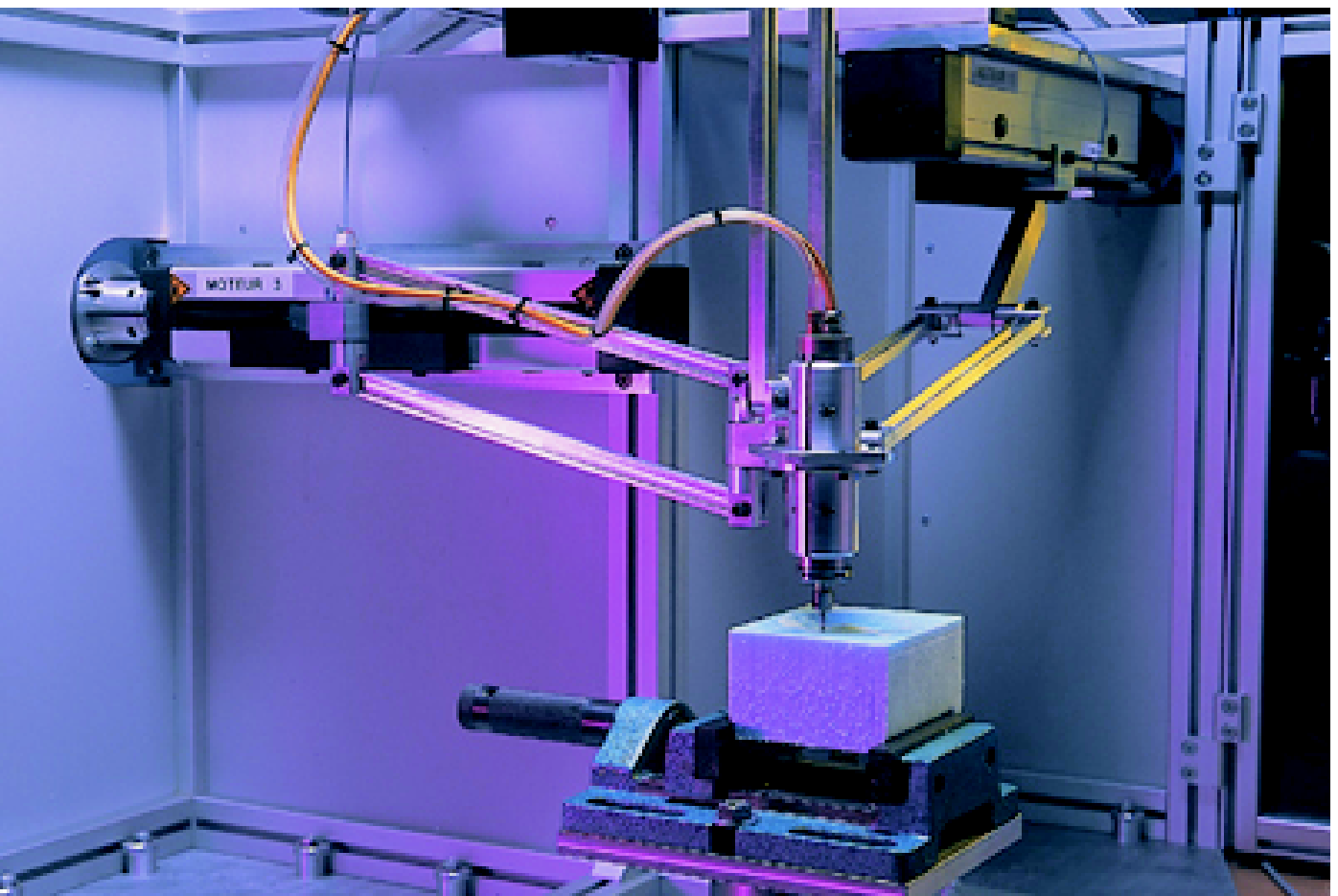,scale = 0.5}}
    \caption{The Orthoglide}
    \label{fig002}
 \end{center}
 \end{minipage}
 \end{figure}
Unlike Lampinen \cite{Lampinen:2003}, who used a genetic
algorithm, or Zhang and Shin \cite{Zhang:2004}, who used what they called a 
relative-motion method, where the relative velocity and
relative acceleration of a follower with respect to a cam is
prescribed and implemented in C++, we use a deterministic method
while taking into account geometric and machining constraints, as
outlined in \cite{Bouzakis:1997}. Unlike Carra \cite{Carra:2004},
who used a negative-radius follower to reduce the pressure angle, we
use a positive-radius follower that permits us to assemble several cams
on the same roller-carrier.

In Section~2, we introduce the relations describing the cam
profile and the mechanism kinematics. In Section~3, we derive
conditions on the design parameters so as to have a fully convex
cam profile, to avoid \textit{undercutting}, and to have a
geometrically feasible mechanism. In Section~4, the pressure
angle, a key performance index of cam mechanisms, is studied in
order to choose the design parameters that give the best
pressure angle distribution, a compromise being made with the
accuracy of the mechanism.
\section{Synthesis of the Planar Cam Mechanism}
Let the $x$-$y$ frame be fixed to the machine frame and the
$u$-$v$ frame be attached to the cam, as depicted in
Fig.~\ref{fig003}. $O_{\!1}$ is the origin of both frames, while
$O_{2}$ is the center of the roller, and $C$ is the contact point
between cam and roller.
 \begin{figure}[htb]
 \begin{minipage}[b]{8cm}
 \begin{center}
   \psfrag{f}{$\bf f$}
   \psfrag{p}{$p$}    \psfrag{e}{$e$}    \psfrag{p}{$p$}    \psfrag{s}{$s$}
   \psfrag{d}{$d$}    \psfrag{x}{$x$}   \psfrag{y}{$y$}    \psfrag{P}{$P$}
   \psfrag{C}{$C$}    \psfrag{mu}{$\mu$}
   \psfrag{u}{$u$}       \psfrag{v}{$v$}
   \psfrag{delta}{$\delta$}
   \psfrag{b2}{$b_2$}   \psfrag{b3}{$b_3$}
   \psfrag{a4}{$a_4$}   \psfrag{psi}{$\psi$}
   \psfrag{theta}{$\theta$}
   \psfrag{O1}{$O_1$}   \psfrag{O2}{$O_2$}
   \centerline{\epsfig{file = 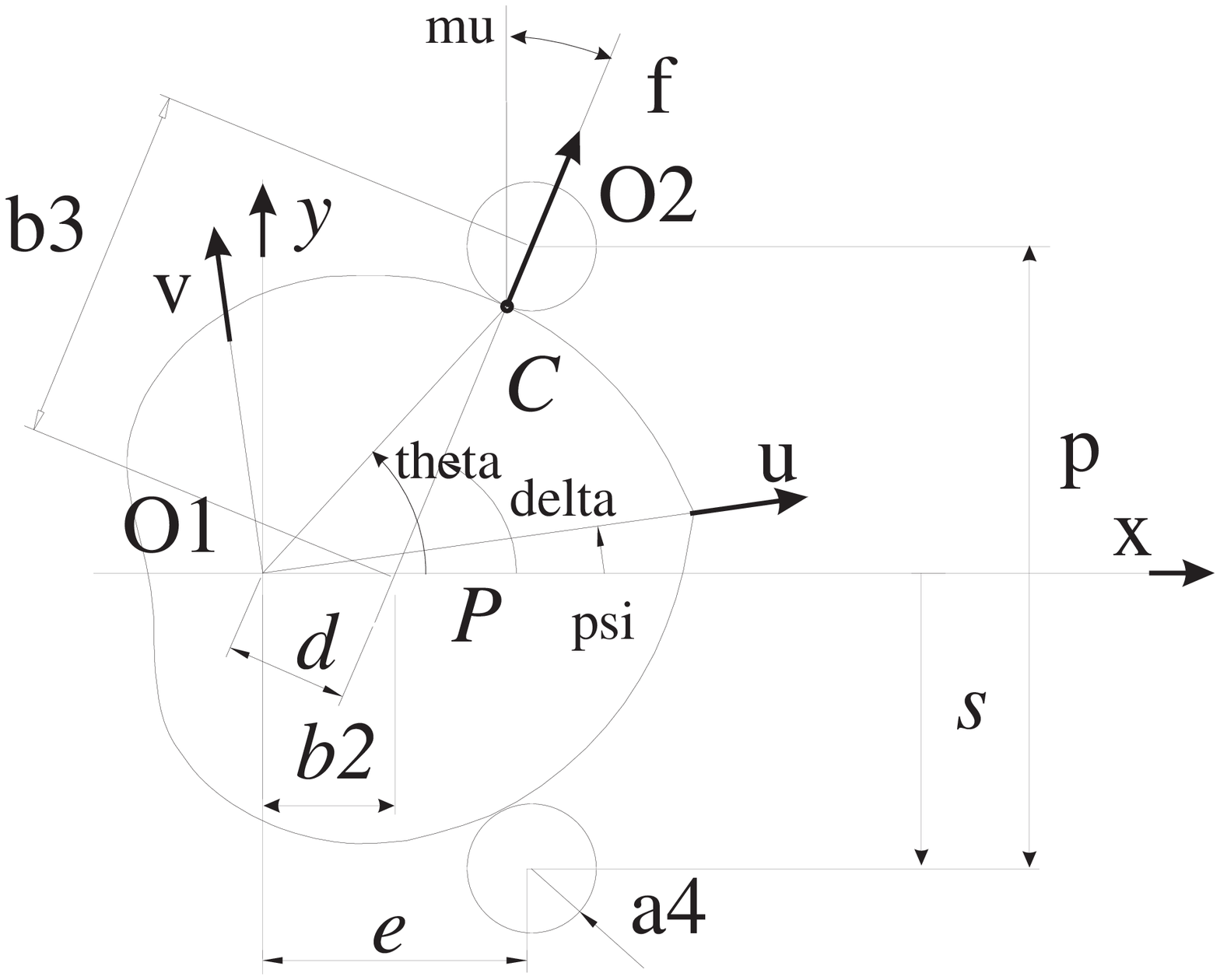,scale = 0.35}}
  \caption{Parameterization of \goodbreak Slide-o-Cam}
  \label{fig003}
 \end{center}
 \end{minipage}
 \begin{minipage}[b]{8cm}
 \begin{center}
   \psfrag{O1}{$O_1$}
   \psfrag{p}{$p$}    \psfrag{x}{$x$}    \psfrag{y}{$y$}
   \psfrag{u}{$u$}    \psfrag{v}{$v$}    \psfrag{x}{$x$}
   \psfrag{s(0)}{$s(0)$}
   \centerline{\epsfig{file = 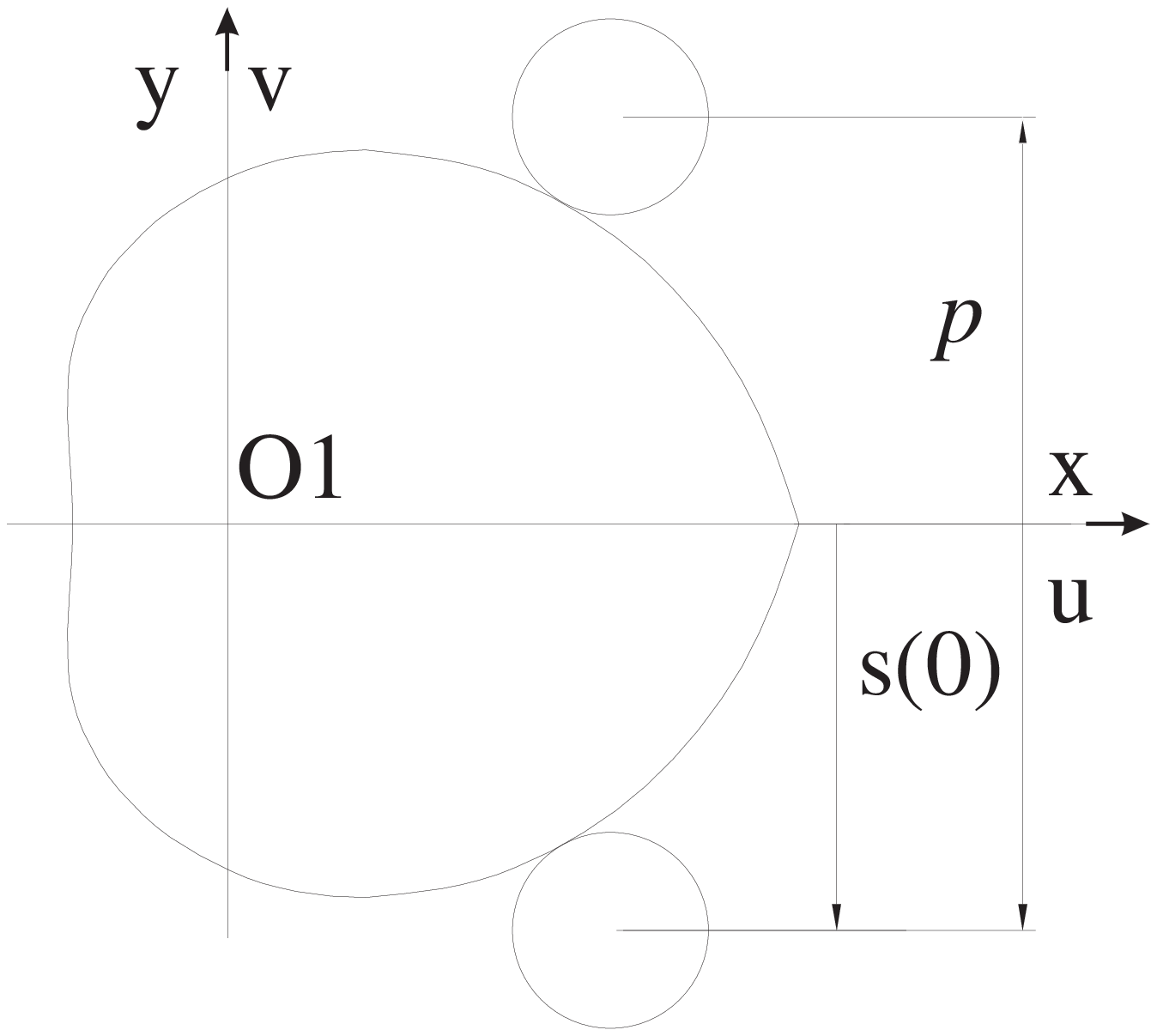,scale = 0.35}}
   \caption{Home configuration of the mechanism} \label{fig004}
 \end{center}
 \end{minipage}
 \end{figure}
The geometric parameters defining the cam mechanism are
illustrated in the same figure. The notation of this figure is
based on the general notation introduced in
\cite{Gonzalez-Palacios:1993}, namely, ($i$) $p$, the pitch, {\it
i.e.}, the distance between the center of two rollers on the same
side of the follower; ($ii$) $e$, distance between the axis of the
cam and the line of centers of the rollers; ($iii$) $a_{4}$,
radius of the roller-bearing, {\it i.e.}, the radius of the
roller; ($iv$) $\psi$, angle of rotation of the cam, the input of
the mechanism; ($v$) $s$, position of the center of the roller,
{\it i.e}, the displacement of the follower,  the output of the
mechanism; ($vi$) $\mu$, pressure angle; ($vii$) {\bf f}, force
transmitted from cam to roller. In this paper, $p$ is set
to 50~mm, in order to meet the Orthoglide specifications.

The above parameters as well as the contact surface on the cam,
are determined by the geometric relations dictated by the
Aronhold-Kennedy Theorem in the plane \cite{Waldron:1999}. When
the cam makes a complete turn ($\Delta \psi=2\pi$), the
displacement of the roller is equal to $p$, the distance between
two rollers on the same side of the roller-carrying slider
($\Delta s =p$). Furthermore, if we consider the home
configuration of the roller as depicted in Fig.~\ref{fig004}, the
roller is below the $x$-axis for $\psi=0$, so that we have
$s(0)=-p/2$. Hence, the input-output function $s$ is
 \begin{equation}
  s(\psi)=\frac{p}{2\pi}\psi-\frac{p}{2} \label{eq01}
 \end{equation}
The expressions for the first and second derivatives of $s(\psi)$
with respect to $\psi$ will be needed:
 \begin{eqnarray}
 s'(\psi)=p/(2\pi) \quad {\rm and} \quad s''(\psi)=0
 \label{eq02}
 \end{eqnarray}
The cam profile is determined by the displacement of the contact
point $C$ around the cam. The Cartesian coordinates of this point
in the $u$-$v$ frame take the form \cite{Gonzalez-Palacios:1993}
 \begin{subequations}
 \begin{eqnarray}
 u_{c}(\psi)\!\!\! &=&\!\!\! ~~b_{2} \cos \psi+(b_{3}-a_{4})\cos(\delta-\psi) \\
 v_{c}(\psi)\!\!\! &=&\!\!\! -b_{2} \sin \psi + (b_{3}-a_{4})\sin(\delta-\psi)
 \end{eqnarray}
 \label{eq04}
 \end{subequations}
 with coefficients $b_{2}$, $b_{3}$ and $\delta$ given by
 \begin{subequations}
 \begin{eqnarray}
 b_{2}  \!\!\!&=&\!\!\! -s'(\psi) \sin \alpha_{1} \\
 b_{3}  \!\!\!&=&\!\!\! \sqrt{(e+s'(\psi)\sin \alpha_{1})^{2}+(s(\psi)
 \sin\alpha_{1})^{2}} \\
 \delta \!\!\!&=&\!\!\! \arctan \left( \frac{-s(\psi) \sin\alpha_{1}}{e+s'(\psi) \sin \alpha_{1}} \right)
 \end{eqnarray}
 \label{eq05}
 \end{subequations}
$\!\!\!$where $\alpha_{1}$ is the directed angle between the axis
of the cam and the translating direction of the follower, positive in the ccw direction. Considering the
orientation adopted for the input angle $\psi$ and for the output
$s$, as depicted in Fig.~\ref{fig003}, we have
 \begin{equation}
 \alpha_{1}=-\pi /2
 \label{eq06}
 \end{equation}
We now introduce the nondimensional design parameter $\eta$, which
will be extensively used:
\begin{equation}
\eta= e/p \label{eq07}
\end{equation}
Thus, from Eqs.~(\ref{eq01}), (\ref{eq02}), (\ref{eq05}a--c),
(\ref{eq06}) and (\ref{eq07}), we find the expressions for
coefficients $b_{2}$, $b_{3}$ and $\delta$ as
 \begin{subequations}
 \begin{eqnarray}
 b_{2}  \!\!\!&=&\!\!\! \frac{p}{2\pi} \\
 b_{3}  \!\!\!&=&\!\!\! \frac{p}{2\pi}\sqrt{(2\pi \eta -1)^{2}+(\psi-\pi)^{2}} \\
 \delta \!\!\!&=&\!\!\! \arctan\left(\frac{\psi-\pi}{2\pi \eta -1} \right)
 \end{eqnarray}
 \label{eq08}
 \end{subequations}
\newline whence a first constraint on $\eta$, $\eta \neq 1/(2\pi)$, is derived.
An {\it extended angle} $\Delta$ is introduced \cite{Lee:2001}, so
that the cam profile closes. Angle $\Delta$ is obtained as a
root of Eq.~\ref{eq04}. In the case of Slide-o-Cam,
$\Delta$ is negative, as shown in Fig.~\ref{fig005}. Consequently,
the cam profile closes within $\Delta \leq \psi \leq 2\pi-\Delta$.
 \begin{figure}[!ht]
 \begin{center}
     \psfrag{O1}{$O_1$}
     \psfrag{x}{$x$}
     \psfrag{y}{$y$}
     \psfrag{u}{$u$}
     \psfrag{v}{$v$}
     \psfrag{C}{$C$}
     \psfrag{Psi}{$\psi$}
     \subfigure[]{\epsfig{file =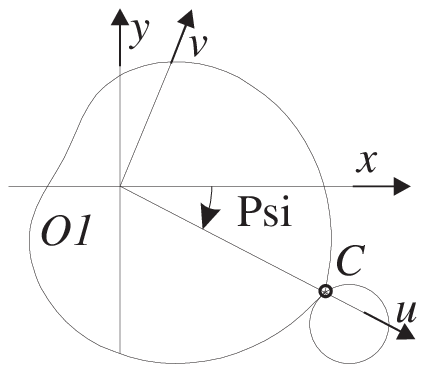,scale = 0.9}}
     \psfrag{x, u}{$x, u$}
     \psfrag{y, v}{$y, v$}
     \psfrag{delta}{$\Delta$}
     \psfrag{-u}{-$u$}
     \psfrag{-v}{-$v$}
     \subfigure[]{\epsfig{file =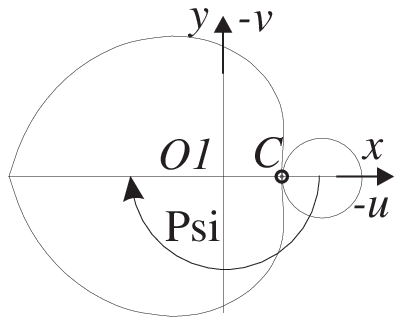,scale = 0.9}}
     \psfrag{u}{$u$}
     \psfrag{v}{$v$}
     \subfigure[]{~~~\epsfig{file =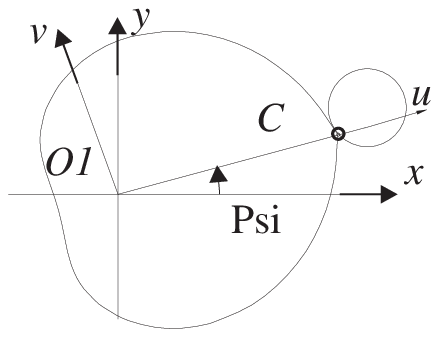,scale = 0.9}}
 \end{center}
 \caption{Orientations of the cam found when $v_c=0$: (a) $\psi=\Delta$; 
  (c) $\psi=\pi$; and (d) $\psi=2\pi-\Delta$}
 \label{fig005}
 \end{figure}
\subsection{Pitch-Curve Determination}
The pitch curve is the trajectory of the center $O_{2}$ of the
roller, distinct from the trajectory of the contact point $C$,
which produces the cam profile. The Cartesian coordinates of point
$O_{2}$ in the $x$-$y$ frame are $(e,s)$, as depicted in
Fig.~\ref{fig003}. Hence, the Cartesian coordinates of the
pitch-curve in the $u$-$v$ frame are
 \begin{subequations}
 \begin{eqnarray}
    u_{p}(\psi) \!\!\!& = \!\!\!& ~~e \cos \psi + s(\psi)\sin \psi
    \label{eq010}\\
    v_{p}(\psi) \!\!\!& = \!\!\!& -e  \sin \psi + s(\psi)\cos \psi
  \end{eqnarray}
 \end{subequations}
\subsection{Geometric Constraints on the Mechanism}
In order to lead to a feasible mechanism, the radius $a_{4}$ of
the roller must satisfy two conditions, as shown in
Fig.~\ref{fig006}a:

\begin{itemize}
\item  Two consecutive rollers on the same side
of the roller-carrying slider must not interfere with each other.
Since $p$ is the distance between the center of two consecutive
rollers, we have the constraint $2a_{4} < p$. Hence the first
condition on $a_{4}$:
     \begin{equation}
     \label{eq011}
     a_{4} / p< 1/2
     \end{equation}
\item  The radius $b$ of the shaft on which the cams are
mounted must be taken into consideration. Hence, we have the
constraint $a_{4}+b \leq e$, the second constraint on $a_{4}$ in
terms of the parameter $\eta$ thus being
\end{itemize}
     \begin{equation}
     \label{eq012}
     a_{4} / p \leq \eta -b / p
     \end{equation}
\begin{figure}[!ht]
\center
  \psfrag{O1}{$O_1$}
  \psfrag{x}{$x$}
  \psfrag{y}{$y$}
  \psfrag{p}{$p$}
  \psfrag{e}{$e$}
  \psfrag{b}{$b$}
  \psfrag{C}{$C$}
  \psfrag{a4}{$a_4$}
  \subfigure[]{\epsfig{file =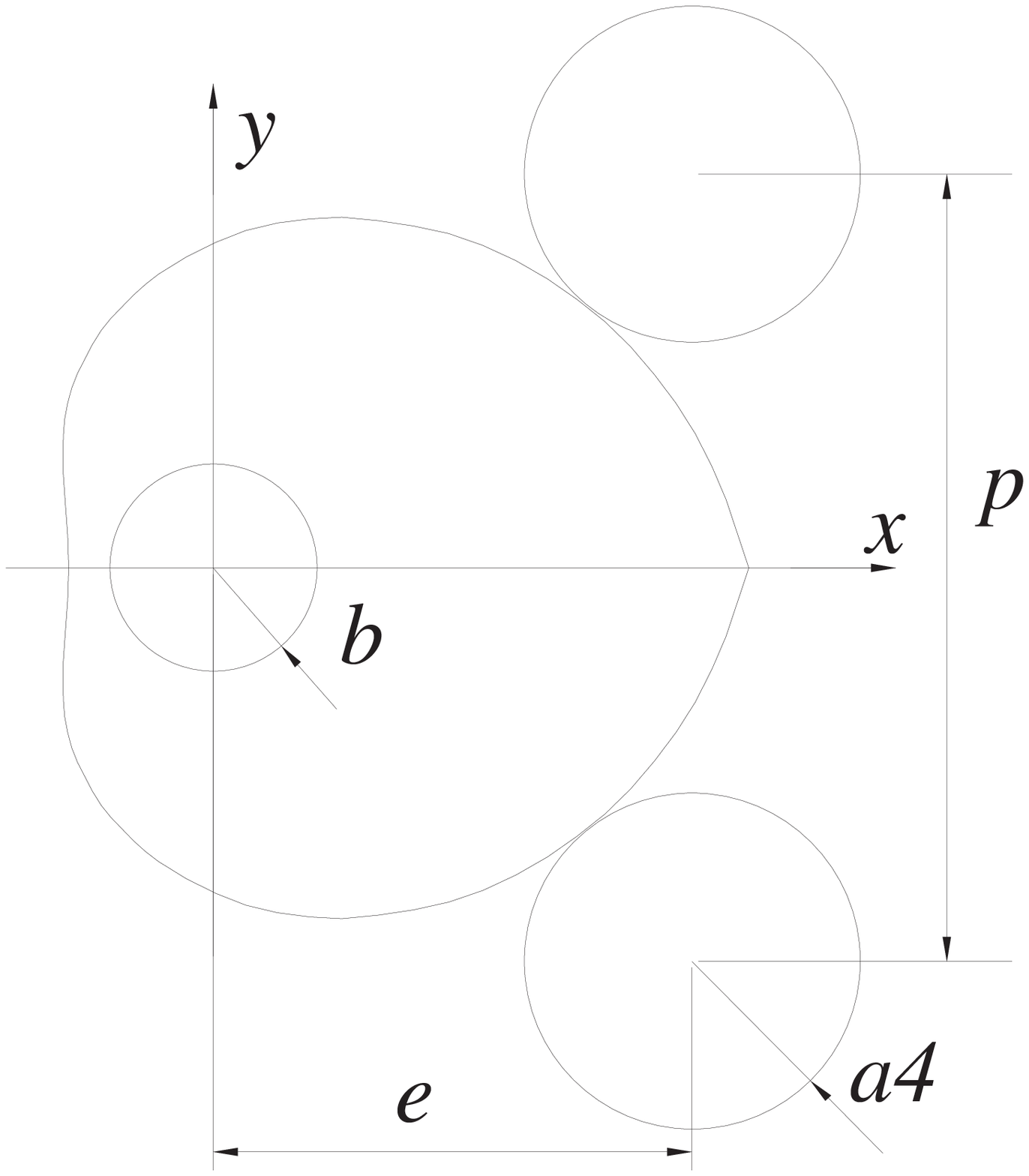,scale = 0.26}~~~~~~~}
  \subfigure[]{~~~~~~~~\epsfig{file =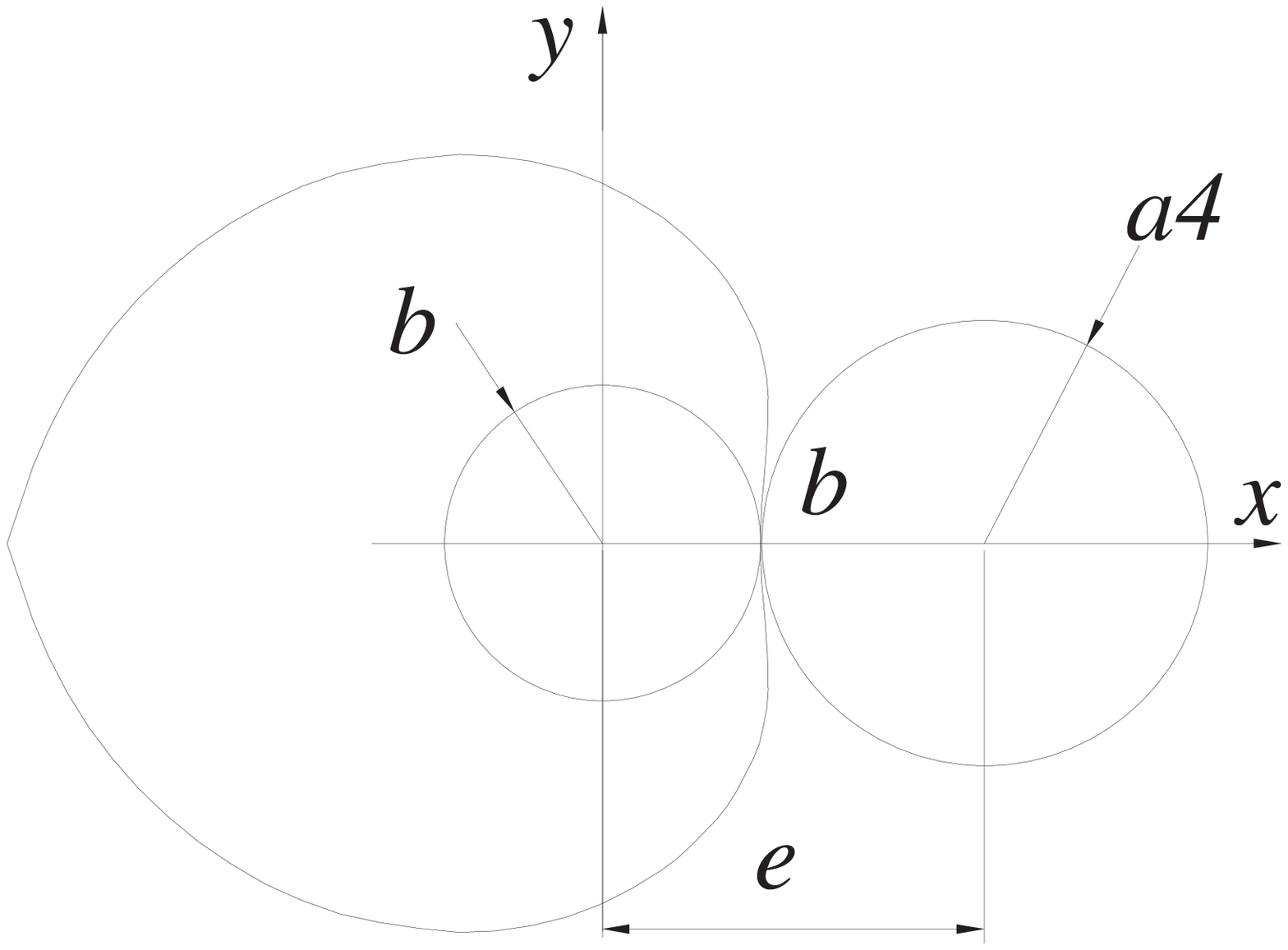,scale = 0.26}}
  \caption{Constraints on the radius of the roller: (a) $ a_{4} / p < 1/2$; and (b) $a_{4}/p \leq \eta -b/p$.}
  \label{fig006}
\end{figure}
Considering the initial configuration of the roller, as depicted
in Fig.~\ref{fig004}, the $v$-component of the Cartesian
coordinate of the contact point $C$ is negative in this
configuration, {\it i.e.}, $ v_{c}(0) \leq 0 $. Moreover, from the
expression for $v_{c}(\psi)$ and for parameters $b_{3}$ and
$\delta$ given in Eqs.~(\ref{eq04}b), (\ref{eq08}b \& c),
respectively, the above relation leads to the condition:
 \beqa
  \left( \frac{p}{2\pi a_4}\sqrt{(2\pi \eta-1)^{2}+(-\pi)^{2}} - 1 \right)
  \sin \left[\arctan\left(\frac{-\pi}{2\pi \eta -1}\right) \right]
  \leq 0 \nonumber
  \eeqa
Further, we define $A$ and $B$ as:
 \beqa
 A=\frac{p}{2\pi a_4}\sqrt{(2\pi \eta-1)^{2}+\pi^{2}} - 1
 ~\mbox{and}~
 B=\sin\left[\arctan\left(\frac{-\pi}{2\pi \eta -1}\right) \right]
 \nonumber
 \eeqa
Since $(2\pi \eta-1)^{2} >0$, we have
 \beqa
    \sqrt{(2\pi \eta-1)^{2}+\pi^{2}} > \pi
 \eeqa
Hence, $ A > p/(2a_4) - 1$. Furthermore, from the constraint on
$a_{4}$, stated in Eq.~(\ref{eq011}), we have $p/(2a_4)-1
> 0$, whence $A>0$. Consequently, the constraint $v_{c}(0) \leq 0$
leads to the constraint $B \leq 0$.
We rewrite the constraint $B$, by using the trigonometric relation, 
\[ \forall x \in \mathbb{R},~ \sin(\arctan x)=\frac{x}{\sqrt{1+x^{2}}} \]
Hence the constraint $v_{c}(0) \leq 0$ becomes
 \beqa
    B \leq 0 \Longleftrightarrow \frac{-\pi}{(2\pi \eta-1) \sqrt{1+\pi^{2}/(2\pi \eta-1)^{2}}} \leq 0
    \nonumber
 \eeqa
which holds only if $2\pi \eta-1 > 0$. Finally, the constraint $v_{c}(0) \leq 0$ leads to a constraint on $\eta$, namely,
\begin{equation}
 \label{eq013}
 \eta> 1/(2\pi)
\end{equation}
\subsection{Pressure Angle}
The pressure angle is defined as the angle between the common
normal at the cam-roller contact point $C$ and the velocity of the
follower \cite{Angeles:1991}, as depicted in Fig.~\ref{fig003},
where the presure angle is denoted by $\mu$. The smaller $|\mu|$,
the better the force transmission. In the case of high-speed
operations, {\it i.e.}, angular velocities of cams exceeding
50~rpm, the recommended bounds of the pressure angle are within
30$^{\circ}$. Nevertheless, as it is not always possible to have a
pressure angle that remains below 30$^{\circ}$, we adopt the
concept of  \textit{service factor}, which is the percentage of
the working cycle with a pressure angle within 30$^{\circ}$
\cite{Lee:2001}. The service angle will be useful to take into
consideration these notions in the ensuing discussion, when
optimizing the mechanism.

For the case at hand, the expression for the pressure angle $\mu$
is given in \cite{Angeles:1991} as
 \[
 \mu=\arctan \left( \frac{s'(\psi)-e}{s(\psi)} \right)
 \]
Considering the expressions for $s$ and $s'$, and using the
parameter $\eta$ given in Eqs.~(\ref{eq01}), (\ref{eq02}a) and
(\ref{eq07}), respectively, the expression for the pressure angle
becomes
 \begin{equation}
 \mu=\arctan \left( \frac{1-2\pi \eta}{\psi - \pi} \right)
 \label{eq0111}
 \end{equation}
We are only interested in the value of the pressure angle with the
cam driving the roller, which happens with
 \begin{equation}
   \label{eq01111} \pi \leq \psi \leq 2\pi-\Delta
 \end{equation}
Indeed, if we start the motion in the home configuration
depicted in Fig.~\ref{fig004}, with the cam rotating in the ccw
direction, the cam begins to drive the roller only when
$\psi=\pi$; and the cam can drive the follower until contact is
lost, {\it i.e.}, when $\psi=2\pi-\Delta$, as shown in
Figs.~\ref{fig005}b \& c.

Nevertheless, as shown in Fig.~\ref{fig008}, the conjugate cam can
also drive the follower when $0 \leq \psi \leq \pi-\Delta$; there
is therefore a common interval, for $\pi \leq \psi \leq
\pi-\Delta$, during which two cams can drive the follower. In this
interval, the conjugate cam can drive a roller with lower absolute
values of the pressure angle. We assume that, when the two cams
can drive the rollers, the cam with the lower absolute value of
pressure angle effectively drives the follower. Consequently, we
are only interested in the value of the pressure angle in the
interval,
 \begin{equation}
 \label{eq0080} \pi-\Delta \leq \psi \leq 2\pi-\Delta
 \end{equation}
We study here the influence of parameters $\eta$ and $a_{4}$ on
the values of the pressure angle while the cam drives the roller,
{\it i.e.}, with $\pi-\Delta \leq \psi \leq 2\pi-\Delta$, as
explained above.
 \begin{itemize}
 \item  {\it Influence of parameter $\eta$}: Figure~\ref{fig007}
shows the influence of the parameter $\eta$ on the pressure angle,
with $a_{4}$ and $p$ being fixed. From these plots we have one
result: {\it The lower $\eta$, the lower $|\mu|$.}

 \item {\it Influence of the radius of the roller $a_{4}$:}
$a_{4}$ does not appear in the expression for the pressure angle,
but it influences the value of the extended angle $\Delta$, and
hence, the plot boundaries of the pressure angle, as shown in
Fig.~\ref{fig008}.
 \end{itemize}

By computing the value of the extended angle $\Delta$ for several
values of $a_{4}$, we can see that the higher $a_{4}$, the lower
$|\Delta|$. Consequently, since the boundaries to plot the
pressure angle are $\pi-\Delta$ and $2\pi-\Delta$, we can see that
when we increase $a_{4}$, $-\Delta$ decreases and the boundaries
are translated toward the left, {\it i.e.}, toward higher absolute
values of the pressure angle.
\begin{figure}[!hb]
 \begin{center}
 \begin{minipage}[b]{8cm}
 \psfrag{mu}{$\mu$}
 \psfrag{psi}{$\psi$}
 \psfrag{pi}{$\pi$}
 \psfrag{pi-Delta}{$\pi-\Delta$}
 \psfrag{2pi-Delta}{$2\pi-\Delta$}
 \psfrag{-6}{-6}
 \psfrag{-2}{-2}
 \psfrag{2}{2}
 \psfrag{6}{6}
 \psfrag{10}{10}
 \psfrag{100}{100}
 \psfrag{50}{50}
 \psfrag{-50}{-50}
 \psfig{file= 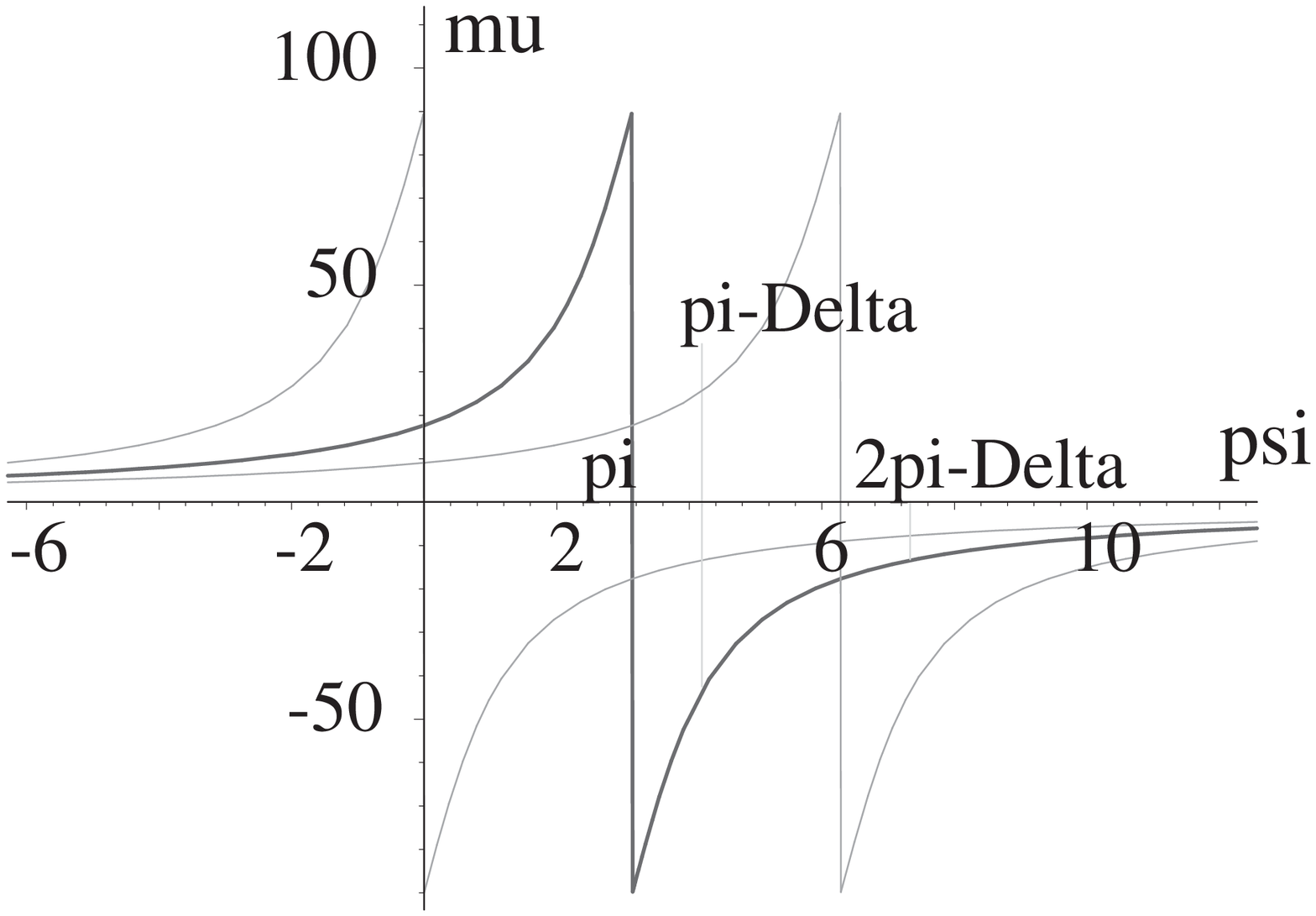, scale= 0.4}
 \caption{Pressure angle distribution}  \label{fig008}
 \end{minipage}
 \begin{minipage}[b]{8cm}
 \psfrag{mu}{$\mu$}
 \psfrag{psi}{$\psi$}
 \psfrag{eta=5}{$\eta=5$}
 \psfrag{eta=2}{$\eta=2$}
 \psfrag{eta=1.5}{$\eta=1.5$}
 \psfrag{eta=1}{$\eta=1$}
 \psfrag{eta=0.8}{$\eta=0.8$}
 \psfrag{eta=0.4}{$\eta=0.4$}
 \psfrag{eta=1/pi}{$\eta=1/\pi$}
 \psfrag{8}{8}  \psfrag{6}{6}  \psfrag{4}{4}  \psfrag{2}{2}
 \psfrag{-20}{-20}  \psfrag{-40}{-40}  \psfrag{-60}{-60}  \psfrag{-80}{-80}
 \psfig{file= 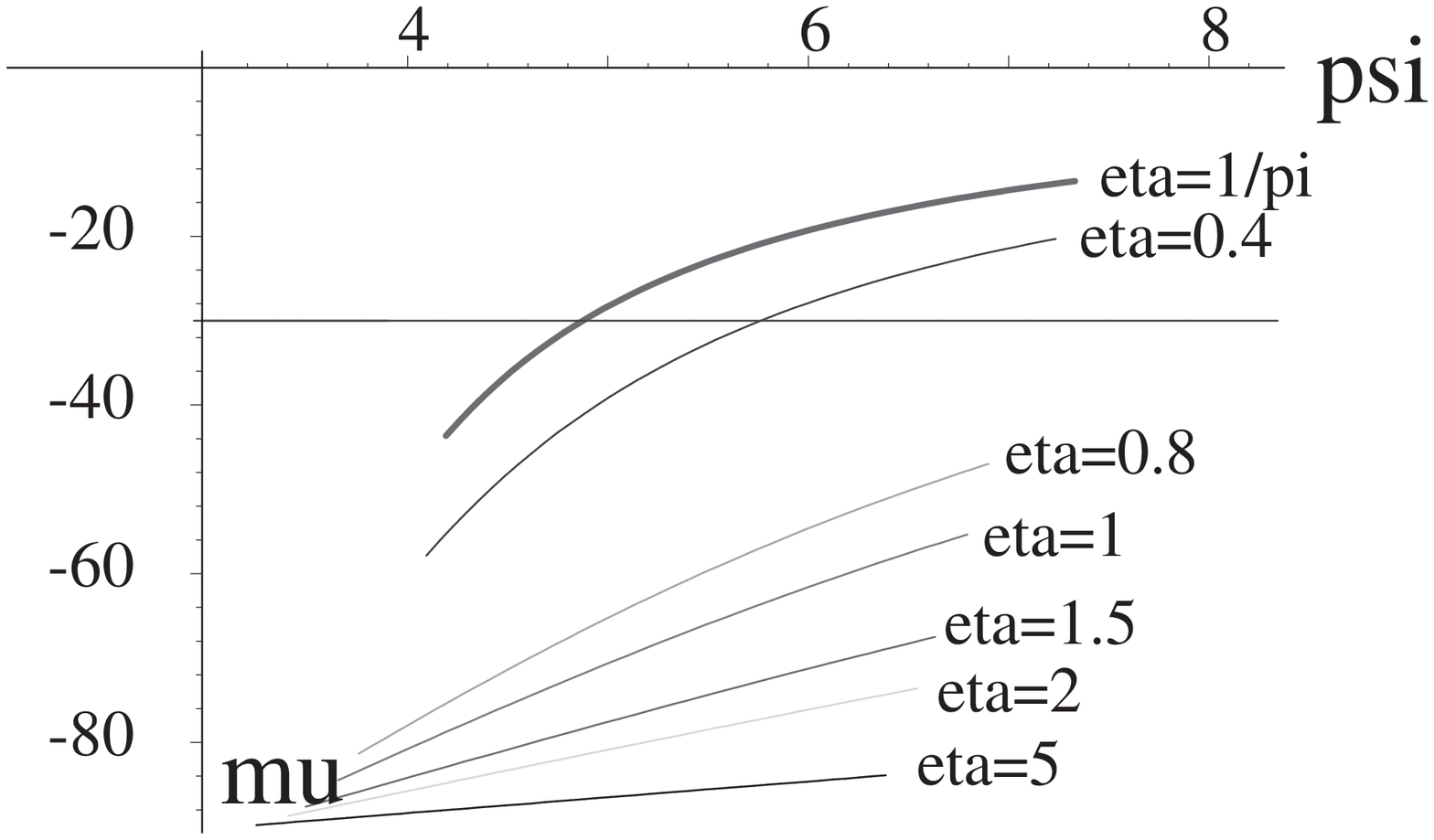, scale= 0.4}
 \caption{Influence of parameter $\eta$ on the pressure angle
 $\mu$ (in degree), with $p=50$~mm and $a_{4}=10$~mm} \label{fig007}
 \end{minipage}
 \end{center}
 \end{figure}
\section{Convexity of the Cam Profile and Undercutting}
In order to enhance machining accuracy, we need the cam profile to
be fully convex. In this section we establish conditions on the
design parameters $\eta$ and $a_{4}$ in order to have a fully
convex cam profile. We thus study the sign-change in the curvature of the
cam profile via that of the pitch curve. Furthermore, for cam
design in roller-follower mechanisms, we should also consider
\textit{undercutting}. Undercutting occurs when the radius of the
roller is greater than or equal to the minimum absolute value of
the radius of curvature of the pitch curve. Upon avoiding
undercutting, the  sign of the curvature of the pitch curve is
identical to that of the cam profile.
\subsection{Curvature of the Cam Profile}
The curvature of any planar parametric curve, in terms of the
Cartesian coordinates $u$ and $v$, and parameterized with any
parameter $\psi$, is given by \cite{Angeles:1991}:
 \begin{equation}
 \label{eq1}
 \kappa=\frac{v'(\psi)u''(\psi)-u'(\psi)v''(\psi)}{[u'(\psi)^{2}+v'(\psi)^{2}]^{3/2}}
 \end{equation}
The sign of $\kappa$ in Eq.~(\ref{eq1}) tells whether the curve is
convex or concave at a point: a positive $\kappa$ implies 
convexity, while a negative $\kappa$ implies concavity at that
point. To obtain the curvature of the cam profile for a given
roller-follower, we use the Cartesian coordinates of the pitch
curve, since obtaining its first and second derivatives leads to
simpler expressions as compared with those associated with the cam
profile itself. Then, the curvature of the cam profile is derived
by a simple geometric relationship between the curvatures of the
pitch curve and of the cam profile.

The Cartesian coordinates of the pitch curve were recalled in
Eqs.~(\ref{eq010} \& b), while Eqs.~(\ref{eq02}a \& b) give their
first and second derivatives as
 \beqa
 s(\psi)= p/(2\pi)\psi-p/2~,~~~
 s'(\psi)  =  p / (2\pi)~,~~~
 s''(\psi)  =  0 \nonumber
 \eeqa
With the above-mentioned expressions, we can compute the first and
second derivatives of the Cartesian coordinates of the pitch curve
with respect to the angle of rotation of the cam, $\psi$:
 \begin{subequations}
 \begin{eqnarray}
 u'_{p}(\psi)  & = & [s'(\psi)-e]\sin \psi + s(\psi)\cos \psi \label{eq2} \\
 v'_{p}(\psi)  & = & [s'(\psi)-e]\cos \psi - s(\psi)\sin \psi  \\
 u''_{p}(\psi) & = & [2s'(\psi)-e]\cos \psi - s(\psi)\sin \psi  \\
 v''_{p}(\psi) & = &-[2s'(\psi)-e]\sin \psi - s(\psi)\cos \psi
 \end{eqnarray}
 \end{subequations}
By substituting $\eta=e/p$, along with Eqs.~(\ref{eq2}--d), into Eq.~(\ref{eq1}), the curvature $\kappa_{p}$ of the pitch curve is obtained as
 \begin{equation}
 \label{eq6}
 \kappa_{p}=\frac{2\pi}{p} \frac{[(\psi-\pi)^{2}+2(2\pi
\eta-1)(\pi \eta-1)]}{[(\psi-\pi)^{2}+(2\pi \eta-1)^{2}]^{3/2}}
 \end{equation}
\noindent which remains bounded by virtue of condition (\ref{eq013}). 

\noindent Let $\rho_{c}$ and $\rho_{p}$ be the radii of curvature of both the cam profile and the pitch curve, respectively, and $\kappa_{c}$ the curvature of the cam profile. Since the curvature is the reciprocal of the radius of curvature, we have $\rho_{c} = 1/\kappa_{c}$ and $\rho_{p} = 1/\kappa_{p}$. Furthermore, due to
the definition of the pitch curve, it is apparent that \begin{equation}  \label{eq8} \rho_{p} = \rho_{c} + a_{4}
 \end{equation}
Writing Eq.~(\ref{eq8}) in terms of $\kappa_{c}$ and $\kappa_{p}$, we obtain the curvature of the cam profile as
 \begin{equation}
 \label{eq9} \kappa_{c}=\frac{\kappa_{p}}{1-a_{4} \kappa_{p}}
 \end{equation}
with $\kappa_{p}$ given in Eq.~(\ref{eq6}). As we saw previously, we want the cam profile to be fully convex, which happens if the pitch curve is fully convex too. We thus find first the convexity condition of the pitch curve.
\subsection{Convexity Condition of the Pitch Curve}
Considering the expression for $\kappa_{p}$ in Eq.~(\ref{eq6}), we have, for every value of $\psi$, $\kappa_{p} \geq 0$ if  $(2\pi \eta-1)(\pi \eta-1) \geq 0$ and $\eta \neq 1 / (2\pi)$, whence the condition on $\eta$:
 \begin{equation}
 \label{eq10}
 \kappa_{p} \geq 0
 ~~~{\rm if}~~~
 \eta \in [0, 1/(2\pi)[ ~~\cup~~ [1/\pi, + \infty[
 \end{equation}
Figure~\ref{fig100} shows pitch-curve profiles and their
curvatures for two values of $\eta$.
\begin{figure}[!ht]
 \center
 \begin{minipage}[b]{8cm}
 \center
 {
  \psfrag{-20}{-$20$}
  \psfrag{-10}{-$10$}
  \psfrag{10}{$10$}
  \psfrag{20}{$20$}
  \psfrag{30}{$30$}
  \psfig{file= 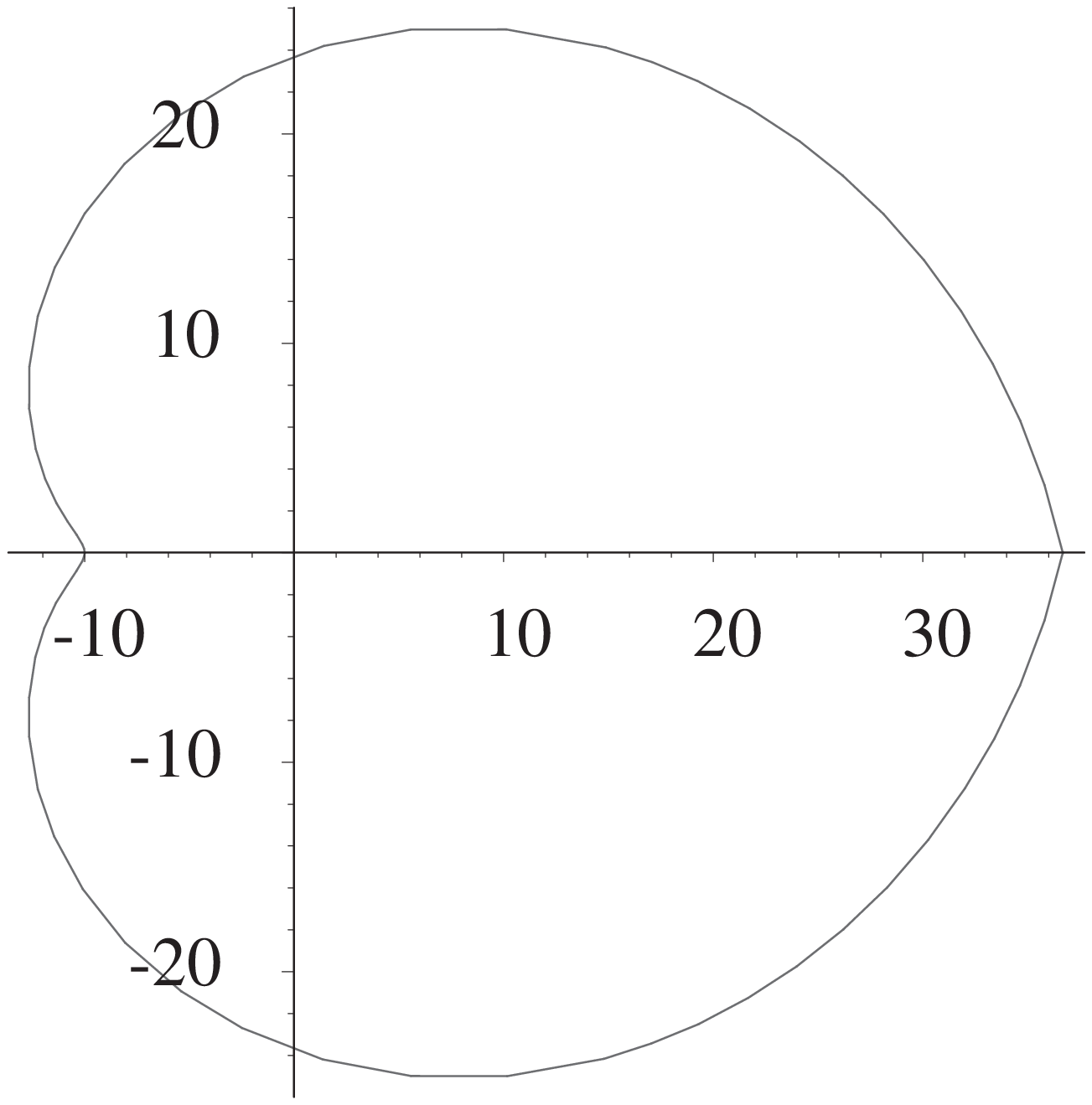,scale=0.25}}
 \end{minipage}
 \begin{minipage}[b]{8cm}
 \center
 {
  \psfrag{-20}{-$20$}
  \psfrag{-40}{-$40$}
  \psfrag{40}{$40$}
  \psfrag{20}{$20$}
  \psfrag{30}{$30$}
  \psfig{file= 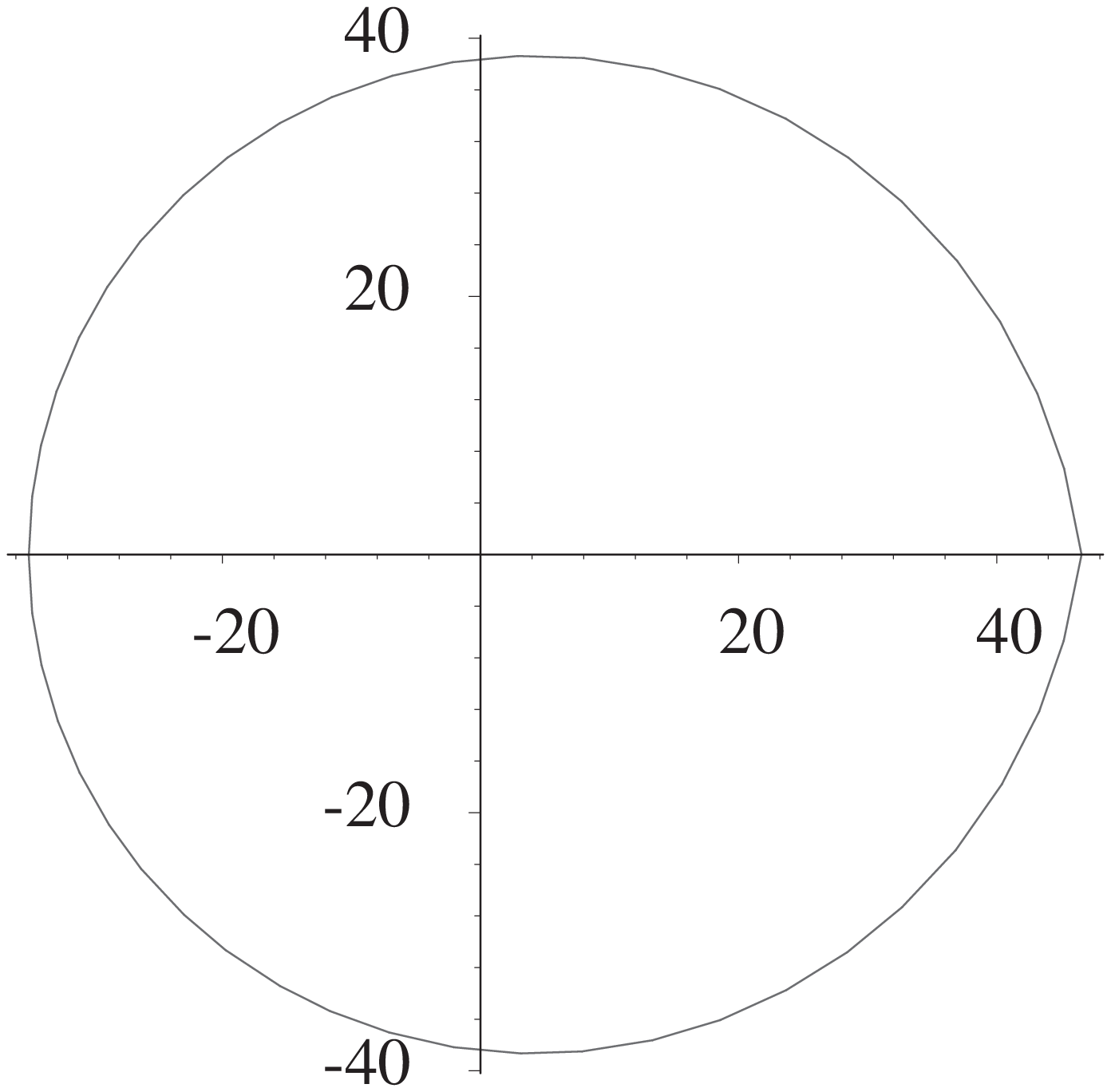,  scale=0.25}}
 \end{minipage}
\par
 \begin{minipage}[b]{8cm}
 \begin{center}
  \psfrag{psi}{$\psi$}
  \psfrag{kappa_p}{$\kappa_p$}
  {
   \psfrag{0.2}{0.2}
   \psfrag{-0.2}{-0.2}
   \psfrag{-0.4}{-0.4}
   \psfrag{-0.6}{-0.6}
   \psfrag{-0.8}{-0.8}
   \psfrag{-1.0}{-1.0}
   \psfrag{-1.2}{-1.2}
   \psfrag{-1.4}{-1.4}
   \psfrag{2}{2}
   \psfrag{4}{4}
   \psfrag{6}{6}
  \subfigure[]
  {~~~~~~~\psfig{file= 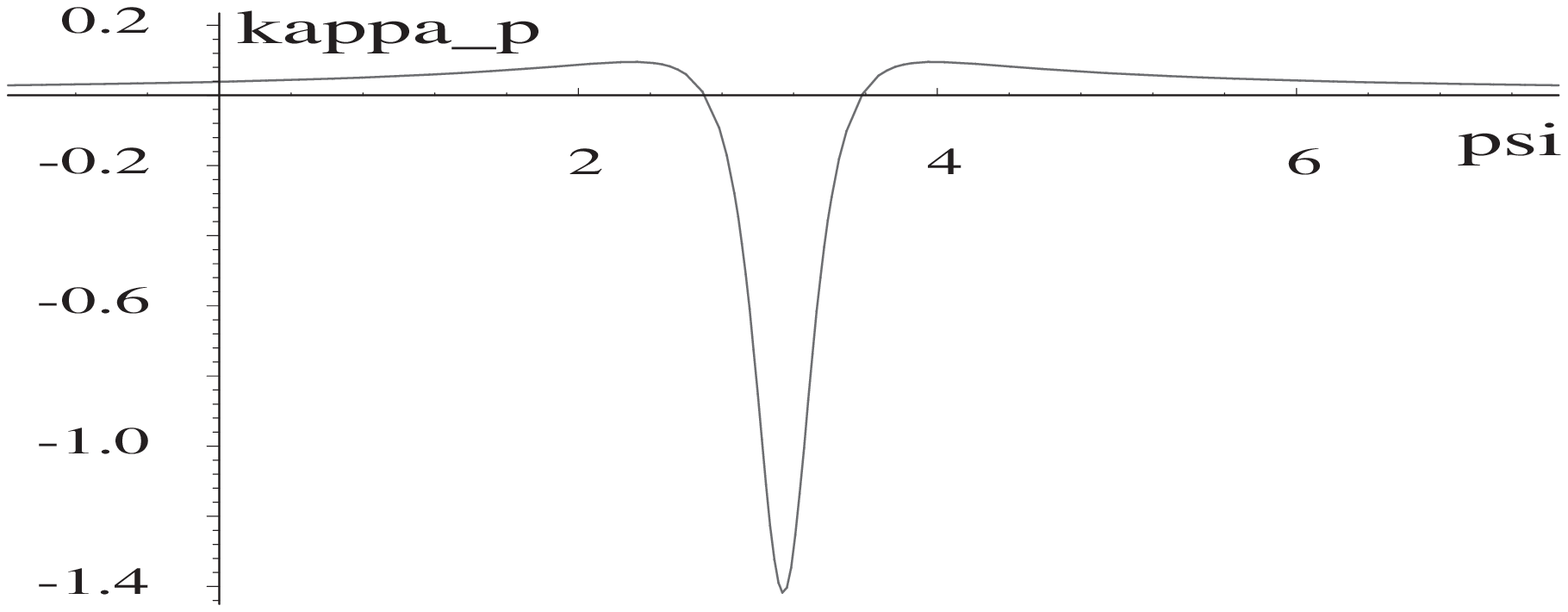, scale= 0.3}~~~~~~~~~~~~~~~}}
 \end{center}
 \end{minipage}
 \begin{minipage}[b]{8cm}
 \begin{center}
  \psfrag{psi}{$\psi$}
  \psfrag{kappa_p}{$\kappa_p$}
  {
   \psfrag{0.022}{0.022}
   \psfrag{0.023}{0.023}
   \psfrag{0.024}{0.024}
   \psfrag{0.025}{0.025}
   \psfrag{0.026}{0.026}
   \psfrag{0}{0}    \psfrag{1}{1}    \psfrag{2}{2}    \psfrag{3}{3}
   \psfrag{4}{4}    \psfrag{5}{5}    \psfrag{6}{6}    \psfrag{7}{7}
  \subfigure[]
  {~~~~~~~~\psfig{file= 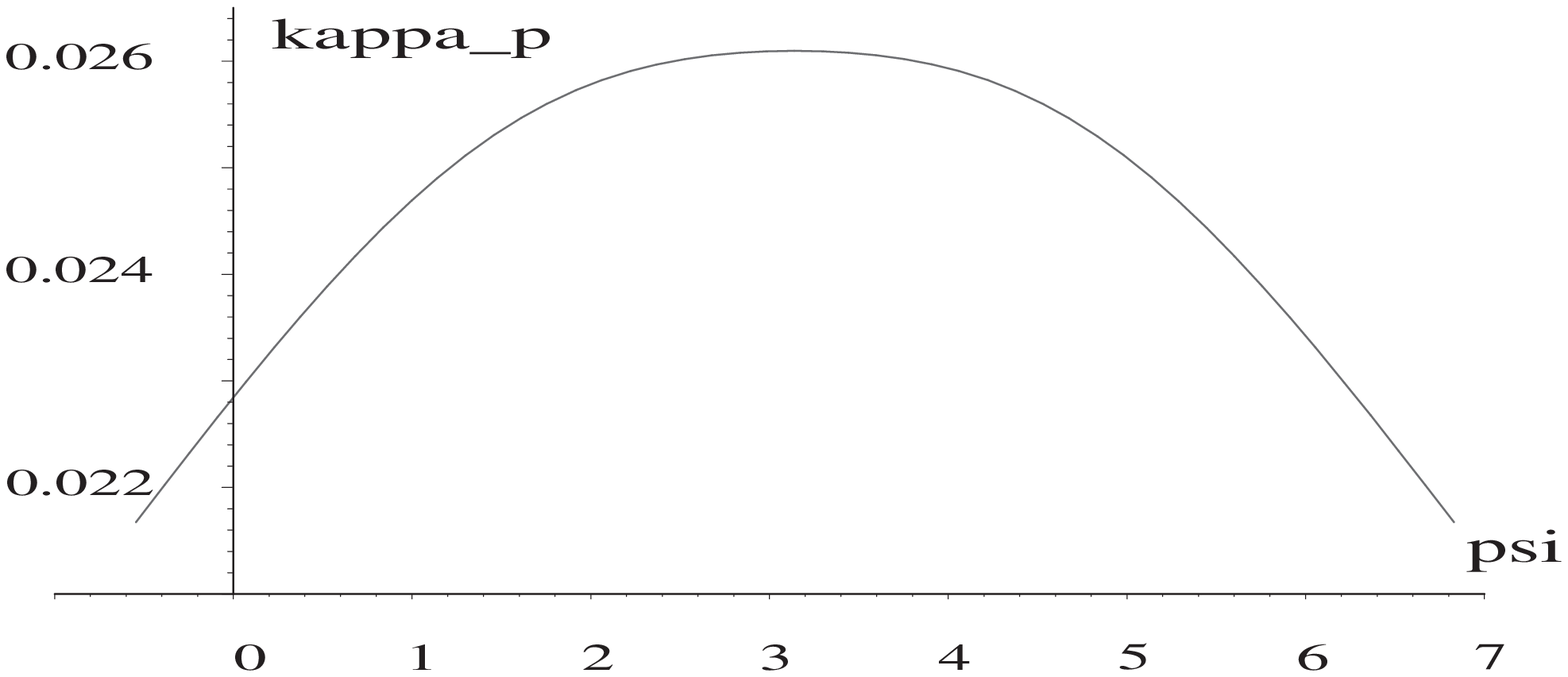, scale=0.3}~~~~~}}
 \end{center}
 \end{minipage}
  \caption{Pitch-curve profiles and their corresponding curvatures with $p=50$~mm: \goodbreak (a) $\eta=0.2$ ($\eta \in \rbrack 1/(2\pi),1/\pi \lbrack $) and (b) $\eta=0.7$ ($\eta>1/\pi$)}
  \label{fig100}
\end{figure}
The condition on $\eta$ given in Eq.~(\ref{eq10}) must be combined
with the condition appearing in Eq.~(\ref{eq013}), $\eta
> 1/(2\pi)$; hence, the final \textbf{convexity condition of the
pitch curve} is:
 \begin{equation}
 \label{eq11}
 \eta \geq 1/\pi
 \end{equation}
\subsection{Undercutting-Avoidance}
We assume in this subsection that the pitch curve is fully convex,
{\it i.e.}, $\kappa_{p} \geq 0$ and $\eta \geq 1/\pi$. In order to
avoid undercutting, {\it i.e.}, in order to have both the cam
profile and the pitch curve fully convex, we need $\kappa_{c}$ to
be positive. Considering the expression for the curvature of the
cam profile $\kappa_{c}$ of Eq.~(\ref{eq9}), the condition to
avoid undercutting is $1-a_{4} \kappa_{p} > 0$, whence the
condition on the radius of the follower $a_{4}$:
 \[
 a_{4} < \frac{1}{\kappa_{p}(\psi)} ~~~~~ \forall ~\psi \in \mathbb{R}
 \]
Since $\kappa_{p}$ is positive, this condition can be written as
 \begin{equation}
 \label{eq12}
 a_{4} < \frac{1}{\kappa_{p\rm max}}
 ~~,\quad \mbox{with} \quad
 \kappa_{p\rm max}= \max_{\psi \in \mathbb{R}} \kappa_{p}(\psi)
 \end{equation}

 \begin{itemize}
 \item {\it Expression for $\kappa_{p\rm max}$}: In order to compute
the expression for $\kappa_{p\rm max}$, we need the first
derivative $\kappa'_{p}$ of $\kappa_{p}$ with respect to $\psi$
and its roots.  With the condition $\eta \geq 1/\pi$, the
expression for $\kappa_{p}$ given in Eq.~(\ref{eq6}) is
differentiable for every value of $\psi$. We thus obtain
 \end{itemize}
 \begin{equation}
 \kappa'_{p}=- \frac{2 \pi}{p} \frac{(\psi-\pi)(\psi^{2}
-2\pi\psi+\pi^{2}+4\eta^{2}\pi^{2}-10\eta\pi+4)}{[(\psi-\pi)^{2}+(2\pi
\eta-1)^{2}]^{5/2}} \nonumber
 \end{equation}
The roots of $\kappa'_{p}$ are, apparently, $\psi_{1}=\pi$ and the
roots $\psi_{2}$ and $\psi_{3}$ of the polynomial
 \[
 P(\psi)=\psi ^{2} -2\pi\psi+\pi^{2}+4\eta^{2}\pi^{2}-10\eta\pi+4
 \]
Let $\beta_{\psi}$ be the discriminant of the equation $P=0$, {\it
i.e.},
 \[
 \beta_{\psi}=-4\eta^{2}\pi^{2}+10\eta\pi-4
 \]
Therefore, the sign of $\beta_{\psi}$ and, consequently, the roots
$\psi_{2}$ and $\psi_{3}$, depend on the value of $\eta$. Let
$\beta_{\eta}$ be the discriminant of $\beta_{\psi}=0$, a
quadratic equation in $\eta$. Hence, $\beta_{\eta}=9\pi^{2}$,
which is positive. The two roots of $\beta_{\psi}$ are $1/2\pi$
and $2/\pi$. Thus,
 \begin{eqnarray*}
 \beta_{\psi}>0 ~~~&\mbox{if}&~~~ \eta \in \left[\frac{1}{\pi},\frac{2}{\pi}\right[
 \quad {\rm or} \quad
 \beta_{\psi}<0 ~~~\mbox{if}~~~ \eta \in \left] \frac{2}{\pi},+\infty \right[ \\
 \beta_{\psi}=0 ~~~&\mbox{if}&~~~ \eta = \frac{2}{\pi}
 \end{eqnarray*}
We now study the roots of $\kappa'_{p}$ according to the value of
$\eta$.

 \begin{itemize}
\item  $\eta \in [ 1/\pi,2/\pi[$: $\beta_{\psi}>0$,
and the polynomial $P$ has two roots $\psi_{2}$ and $\psi_{3}$, so
that $\kappa'_{p}$ has three roots:
     \begin{subequations}
     \begin{eqnarray}
     \psi_{1} &=& \pi \\
     \psi_{2} &=& \pi + \sqrt{-4\eta^{2}\pi^{2}+10\eta\pi-4} \\
     \psi_{3} &=& \pi - \sqrt{-4\eta^{2}\pi^{2}+10\eta\pi-4}
     \end{eqnarray}
     \label{eq14}
     \end{subequations}
 \item  $\eta \in ] 2/\pi,+\infty[$: $\beta_{\psi}<0$,
and the polynomial $P$ has no real roots, so that $\kappa'_{p}$
has only one root, $\psi_{1}=\pi$.
 \item $\eta=2/\pi$: $\beta_{\psi}=0$, and the polynomial $P$ has one
double root equal to $\pi$, so that $\kappa'_{p}$ has one triple
root $\psi_{1}=\pi$.
 \end{itemize}

To decide whether these roots correspond to minima or maxima, we
need to know the sign of the second derivative $\kappa''_{p}$ of
$\kappa_{p}$ with respect to $\psi$, for the corresponding values
of $\psi$. If the second derivative is negative, the stationary
value is a maximum; if positive, a minimum. The expressions for
the second derivatives, as derived with computer algebra, for the values of
$\psi$ given in Eqs.~(\ref{eq14}a--c):
\begin{eqnarray*}
 \kappa''_{p}(\psi_{1}) = \frac{4\pi (\eta\pi -2)}{p |2\eta\pi-1|^{3}(2\eta\pi-1)}
 \quad \quad
 \kappa''_{p}(\psi_{2})=\kappa''_{p}(\psi_{3}) = \frac{8\pi (\eta\pi -2)}{9p(2\eta\pi-1) \sqrt{6\eta\pi-3}}
\end{eqnarray*}

 \begin{itemize}
 \item  If $\eta \in [ 1/\pi,2/\pi[$,
$\kappa''_{p}(\psi_{1})>0$ and
$\kappa''_{p}(\psi_{2})=\kappa''_{p}(\psi_{3})<0$, the curvature
of the pitch curve has one local minimum for $\psi_{1}$ and two
maxima, for $\psi_{2}$ and $\psi_{3}$. Hence, the value of
$\kappa_{p \rm max}$ is
   \begin{equation}
   \label{eq19}
   \kappa_{p \rm max1}=\kappa_{p}(\psi_{2})=\kappa_{p}(\psi_{3})=\frac{4\pi}{3p \sqrt{6\eta\pi-3}}
   \end{equation}
Figure~\ref{fig101}a shows a plot of the pitch-curve curvature
with $\eta=1/\pi$ and $p=50$~mm. Since $\eta$ is taken equal to
the convexity limit $1/\pi$, the curvature remains positive and
only vanishes for $\psi=\pi$.

\item If $\eta \in ] 2/\pi,+\infty[$, $\kappa''_{p}(\psi_{1})<0$,
the curvature of the pitch curve has a maximum for $\psi_{1}$.
Hence, the value of $\kappa_{p \rm max}$ is
   \begin{equation}
   \label{eq20}
   \kappa_{p \rm max2}=\kappa_{p}(\psi_{1})=\frac{4\pi}{p}\frac{
   (2\eta^{2}\pi^{2}-3\eta\pi+1)}{(4\eta^{2}\pi^{2}-4\eta\pi+1)^{3/2}}
   \end{equation}
Figure~\ref{fig100}d shows a plot of the pitch-curve curvature
with $\eta=0.7$ ($\eta>2/\pi$) and $p=50$~mm.

\item If $\eta=2/\pi$, $\kappa''_{p}(\psi_{1})=0$, we cannot
tell whether we are in the presence of a maximum or a minimum. We
solve this uncertainty graphically, by plotting the curvature of
the pitch curve for $\eta=2/\pi$ and $p=50$~mm.
Figure~\ref{fig101}b reveals that the curvature has a maximum for
$\psi_{1}$. The value of this maximum can be obtained by
substituting $\eta$ by $2/\pi$ into either $\kappa_{p \rm max1}$
or $\kappa_{p \rm max2}$, expressed in Eqs.~(\ref{eq19}) and
(\ref{eq20}), respectively.
 \end{itemize}
\begin{figure}
 \begin{center}
 \begin{minipage}[b]{8cm}
  \begin{center}
   \psfrag{kappa_p}{$\kappa_p$}
   \psfrag{psi}{$\psi$ (rad)}
   {
   \psfrag{2}{2}
   \psfrag{4}{6}
   \psfrag{6}{6}
   \psfrag{0.02}{0.02}
   \psfrag{0.04}{0.04}
   \subfigure[]{\psfig{file= 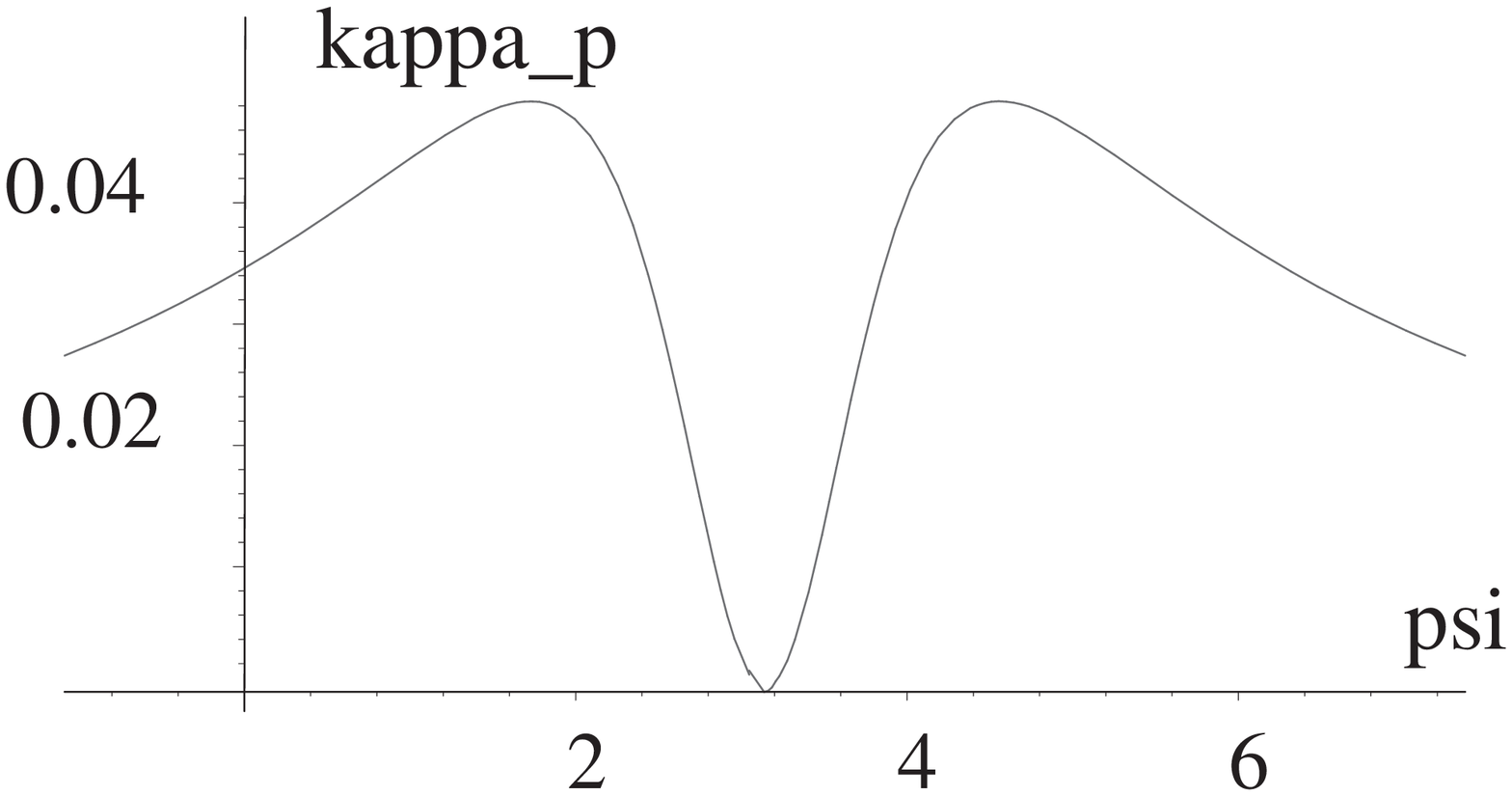, scale=0.3}}}
   \end{center}
 \end{minipage}
 \begin{minipage}[b]{8cm}
   \psfrag{kappa_p}{$\kappa_p$}
   \psfrag{psi}{$\psi$ (rad)}
   {
   \psfrag{0.023}{0.023}
   \psfrag{0.025}{0.025}
   \psfrag{0.027}{0.027}
   \psfrag{0}{0}
   \psfrag{4}{6}
   \psfrag{6}{6}
    \subfigure[]{\psfig{file= 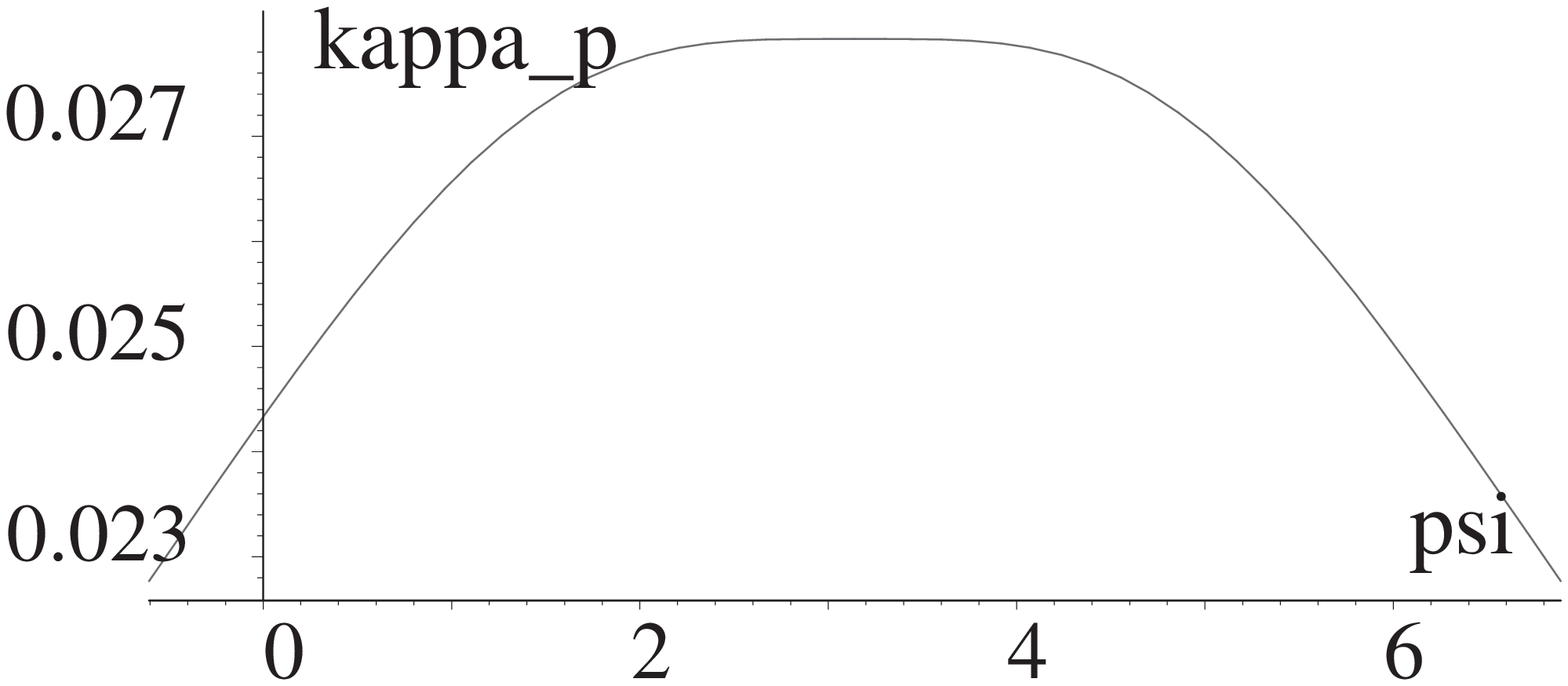, scale=0.3}}}
 \end{minipage}
 \caption{Pitch-curve curvature for $p=50$~mm: (a) $\eta=1/\pi$ and (b) $\eta=2/\pi$} \label{fig101}
 \end{center}
\end{figure}
In summary, to have a fully convex cam profile, taking the
geometric constraints on the mechanism into consideration,
parameter $\eta$ must obey the condition given in
Eq.~(\ref{eq11}), {\it i.e.} $\eta \geq 1/\pi$. We combine the
condition on $a_{4}$ to avoid undercutting, as given in
Eq.~(\ref{eq12}), with the geometric constraints on the mechanism,
as given in Eqs.~(\ref{eq011}) and (\ref{eq012}), which are,
respectively, $a_{4} / p < 1/2$ and $a_{4} / p \leq \eta -b/p$:
 \begin{subequations}
 \begin{eqnarray}
 \mbox{if}~~ \eta \in \left[ \frac{1}{\pi},\frac{2}{\pi} \right], ~~ a_{4} &<&
\min \left\{ \frac{1}{\kappa_{p \rm max1}},~\frac{p}{2},~\eta p-b
\right\} \\
 \mbox{if}~ \eta \in \left[ \frac{2}{\pi},+\infty \right[, ~~ a_{4}
&<& \min \left\{ \frac{1}{\kappa_{p \rm max2}},~\frac{p}{2},~\eta
p-b \right\}
 \end{eqnarray}
 \label{eq21}
 \end{subequations}
where $\kappa_{p \rm max1}$ and $\kappa_{p \rm max2}$ are given in
Eqs.~(\ref{eq19}) and (\ref{eq20}), respectively.
\section{Optimization of the Roller Pin}
We concluded in the previous section that the lowest values of
parameters $\eta$ and $a_{4}$ led to the lowest values of the
pressure angle. Nevertheless, we must take into consideration that
the smaller the radius of the roller, the bigger the deformation
of the roller pin, and hence, a decrease of the stiffness and the
accuracy of the mechanism. In this section we formulate and solve
an optimization problem to find the best compromise on parameters
$\eta$ and $a_{4}$ to obtain the lowest pressure angle values with
an acceptable deformation of the roller pin. In this section we
adopt the approach proposed by Teng and Angeles \cite{Teng:2003}.
\subsection{Minimization of the Elastic Deformation on the Roller Pins}
Here we find the expression for the maximum elastic deformation on
the pin, which will be minimized under given constraints.
Figure~\ref{fig400} displays the free part of the pin, {\it i.e.},
the part not fixed to the roller-carrying slider, as a cantilever
beam, where the load $F=\sqrt{f_{x}^{2}+f_{y}^{2}}$ denotes the
magnitude of the force {\bf f} transmitted by the cam. This force
is applied at a single point at the end of the pin in the worst
loading case. Although the dimensions of the pin are not those of
a simple beam, we assume below that the pin can be modeled as
such, in order to obtain an explicit formula for its deflection.
This assumption was found to be plausible by testing it with FEA
\cite{Teng:2003}.
 \begin{figure}[htb]
  \begin{center}
   \psfrag{f}{$\bf f$}
   \psfrag{L}{$L$}
   \psfrag{a5}{$a_5$}
   \psfrag{x}{$x$}
   \psfrag{y}{$y$}
   \psfrag{z}{$z$}
   {\epsfig{file = 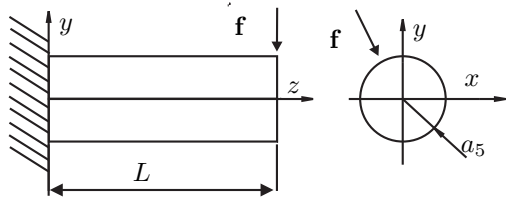,scale = 0.4}}
  \caption{Approximation of the roller pin as a cantilever beam}
  \label{fig400}
  \end{center}
 \end{figure}

The displacement $v_{L}$ at the free end of the pin turns out to
be
 \begin{equation}
 \label{eq400}
 v_{L}=\sqrt{v_{x}^{2}+v_{y}^{2}}=\frac{FL^{3}}{3EI}
 \end{equation}
where $E$ is the Young modulus and $I=\pi a_{5}^{4}/4$, with
$a_{5}$ denoting the radius of the pin, is the polar moment of
inertia of the cross section. Moreover, $v_{x}$ and $v_{y}$ are
the pin elastic displacements in the $x$- and $y$-directions,
respectively, at the free end.

Before proceeding, we prove that the vertical component $f_{y}$ of
the transmitted force is constant, and hence, we will consider
only the magnitude of the $x$-component of $v_{L}$. Since we
assume that the mechanism undergoes a pure-rolling motion, the
force exerted by the cam onto the roller, denoted by ${\bf
f}=[f_{x}~~f_{y}]^{T}$, passes through the center of the roller,
{\it i.e.}, its line of action passes through points $O_{2}$ and
$C$, as depicted in Fig.~\ref{fig003}. With a constant torque
$\tau$ provided by the motor, we have $\tau=||{\bf f}||d$, where
$d$ denotes the distance from the center of the input axis to the
line of action of the force {\bf f}. Moreover, we have $d=b_{2}
\sin \delta$. Hence, $\tau=||{\bf f}||(b_{2} \sin
\delta)=b_{2}(||{\bf f}|| \sin\delta)=b_{2}f_{y} $. Finally, since
$b_{2}=2\pi/p$ we obtain the expression for $f_{y}$ sought:
 \begin{equation}
  \label{eq404} f_{y}=\frac{2\pi\tau}{p}=F_{0}
 \end{equation}
Since $\tau$ is constant, $f_{y}$ is also constant throughout one
full turn of the cam. Consequently we only have to consider the $x$-component of
\negr f, and hence, $v_{x}$, as given below, should be minimized:
\begin{equation}
\label{eq405}
v_{x}=\frac{|f_{x}|L^{3}}{3EI}=\frac{4|f_{x}|L^{3}}{3E\pi
a_{5}^{4}}=\beta
\frac{|f_{x}|}{a_{5}^{4}},~~~\beta=\frac{4L^{3}}{3E\pi}
\end{equation}
$\beta$ thus being a constant factor. The objective function $z$,
to be minimized, is thus defined as
\begin{equation}
\label{eq406} z=\frac{f_{\rm
max}^{2}}{a_{5}^{4}}~~\rightarrow~~\min_{\eta,a_{4},a_{5}}
\end{equation}
where $f_{\rm max}$ is the maximum value of $f_{x}$ throughout one
cycle. Since
\begin{equation}
\label{eq407}
f_{x}=\frac{f_{y}}{\tan\delta}=\frac{F_{0}}{\tan\delta}
\end{equation}
we obtain
 \beqa
z=\frac{1}{a_{5}^{4}} \max_{\psi} \left\{ f_{x}^{2} \right\}
=\frac{1}{a_{5}^{4}} \max_{\psi} \left\{
\frac{F_{0}^{2}}{\tan^{2}\delta} \right \}
=\frac{F_{0}^{2}}{a_{5}^{4}} \max_{\psi} \left\{
\frac{1}{\tan^{2}\delta} \right \} \nonumber
 \eeqa
with $\delta$, a function of $\psi$, given in Eq.~(\ref{eq08}c).
Moreover, the system operates by means of two conjugate
mechanisms, which alternately take over the power transmission. We
established in Eq.~(\ref{eq0080}) that when one mechanism is in
positive action, $\psi$ is bounded between $\psi_{i}=\pi-\Delta$
and $\psi_{f}=2\pi-\Delta$, which corresponds to $\delta$ bounded
between $\delta_{i}$ and $\delta_{f}$ with $0 \leq \delta_{i} <
\delta_{f} \leq \pi$. Moreover, functions $1/\tan^{2}\delta$ and
$\cos^{2}\delta$ are both unimodal in $-\pi \leq \delta \leq 0$
and in $0 \leq \delta \leq \pi$, their common maxima finding
themselves at $-\pi$, $0$ and $\pi$. Since $\cos^{2}\delta$ is
better behaved than $1/\tan^{2}\delta$, we redefine $z$ as
 \beqa
   z=\frac{1}{a_{5}^{4}} \max_{\delta_{i} \leq \delta \leq
\delta_{f}} \left\{ \cos^{2}\delta \right \} \nonumber
  \eeqa
Furthermore, the function $\cos^{2}\delta$ attains its global
minimum of $0$ in $[0,~\pi]$, its maximum in the interval
$[\delta_{i},~\delta_{f}]$, included in $[0,~\pi]$, occurring at
the larger of the two extremes of the interval, $\delta_{i}$ or
$\delta_{f}$. It follows that the objective function to be
minimized becomes
 \beqa
  z=\frac{1}{a_{5}^{4}} \max \left\{
\cos^{2}\delta_{i},~\cos^{2}\delta_{f} \right \} \longrightarrow
\min_{\nu, a_4, a_5} \nonumber
  \eeqa
with $\delta_{i}$ and $\delta_{f}$ the values of $\delta$ for
$\psi_{i}=\pi-\Delta$ and $\psi_{f}=2\pi-\Delta$, respectively.
Using the expression for $\delta$ given in Eq.~(\ref{eq08}c) and
the trigonometric identity
\[ \cos (\arctan x)=\frac{1}{\sqrt{1+x^{2}}} \]
we obtain the expression for $\cos^{2}\delta$:
 \beqa
   \cos^{2}\delta=\frac{(2\pi \eta-1)^{2}}{(2\pi
\eta-1)^{2}+(\psi-\pi)^{2}}
  \eeqa
Hence,
 \begin{subequations}
 \begin{eqnarray}
 \cos^{2}\delta_{i} &=&
 \frac{(2\pi \eta-1)^{2}}{(2\pi\eta-1)^{2}+(\psi_{i}-\pi)^{2}}  \label{eq412}\\
 \cos^{2}\delta_{f} &=& \frac{(2\pi \eta-1)^{2}}{(2\pi\eta-1)^{2}+(\psi_{f}-\pi)^{2}}
 \end{eqnarray}
 \end{subequations}
Furthermore, since $\psi_{i}=\pi-\Delta$, $\psi_{f}=2\pi-\Delta$
and $\Delta<0$, we have $\psi_{f} > \psi_{i} > \pi$, $
\psi_{f}-\pi > \psi_{i}-\pi > 0$ and, consequently, from
Eqs.~(\ref{eq412} \& b), $\cos^{2}\delta_{i}>\cos^{2}\delta_{f}$
and the objective function to minimize becomes
 \begin{equation}
 \label{eq414}
 z=\frac{\cos^{2}\delta_{i}}{a_{5}^{4}}
 \quad  \rightarrow \quad
 \min_{\eta,a_{4},a_{5}}
\end{equation}
with $\cos^{2}\delta_{i}$ given in Eq.~(\ref{eq412}) and
$\psi_{i}=\pi-\Delta$.
\subsection{Geometric Constraints}
Two neighboring pins cannot be tangent to each other, as depicted
in Fig.~\ref{fig401}, and hence the radius $a_{5}$ of the pin is
bounded as
 \be
    a_{5} / p < 1/4
 \label{eq4140}
 \ee
Furthermore, $a_{4}$ and $a_{5}$ are not independent. From the SKF
catalogue, for example, we have information on bearings available
in terms of the outer radius $D$ and the inner radius $d$, as
shown in Fig.~\ref{fig402}. We divide these bearings into five
series, from 1 to 5. Hence, each series can be represented by a
continuous function. We chose series 2, in which the basic dynamic
load rating $C$ lies between 844 and $7020$~N. Furthermore, for
series 2, the relation between $D$ and $d$ can be approximated by
a linear function $D$ vs. $d$, $D \simeq 1.6d+10$ (in mm). Since
$D=2a_{4}$ and $d=2a_{5}$, the above equation leads to
\begin{equation}
\label{eq416} a_{4} \simeq 1.6 a_{5}+5 ~~~\mbox{in mm}
\end{equation}
 \begin{figure}
  \begin{center}
  \begin{minipage}[b]{6cm}
   \begin{center}
   \psfrag{p}{$p$}
   \psfrag{p/2}{$p/2$}
   \psfrag{a4}{$a_4$}
   \psfrag{a5}{$a_5$}
   \psfig{file = 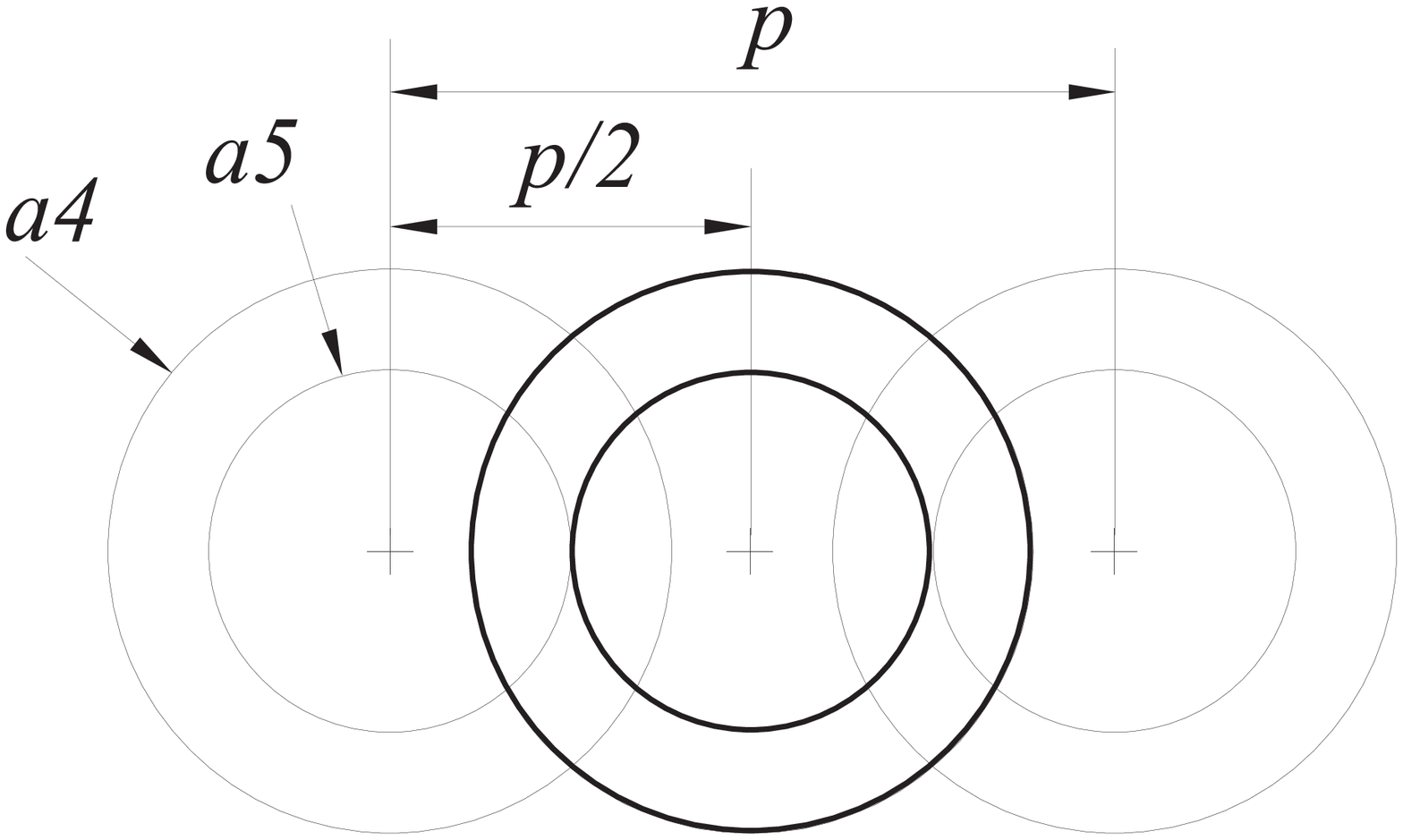,scale = 0.2}
   \caption{Geometric constraint on the roller-pin radius}
   \label{fig401}
   \end{center}
  \end{minipage}
  \begin{minipage}[b]{9cm}
  \psfrag{10}{10}\psfrag{20}{20}\psfrag{30}{30}\psfrag{40}{40}
  \psfrag{50}{50}\psfrag{60}{60}\psfrag{70}{70}\psfrag{0}{0}
  \psfrag{15}{15}\psfrag{5}{5}
  \psfrag{Series 1}{Series 1}   \psfrag{Series 2}{Series 2}   \psfrag{Series 3}{Series 3}
  \psfrag{Series 4}{Series 4}   \psfrag{Series 5}{Series 5}
  \psfrag{d (mm)}{$d$ (mm)}     \psfrag{D (mm)}{$D$ (mm)}
  \psfig{file= 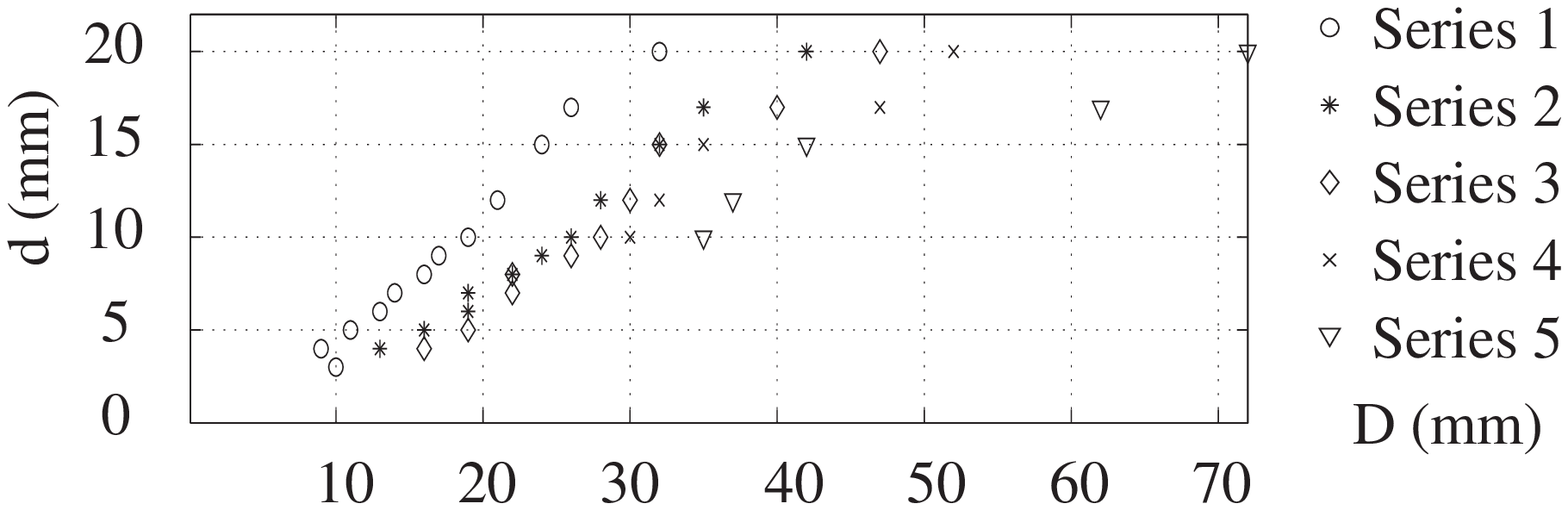, scale= 0.45}
  \caption{Dimensions of the SKF bearings}
  \label{fig402}
  \end{minipage}
   \end{center}
 \end{figure}

\noindent We define now two non-dimensional parameters
$\alpha_{4}$ and $\alpha_{5}$:
 \begin{equation}
 \label{eq417}
 \alpha_{4}= a_{4}/p \quad~~\alpha_{5}=a_{5}/ p
 \end{equation}
where $\alpha_{5}$ can be derived from Eq.~(\ref{eq416}) as
 \begin{equation}
 \label{eq418} \alpha_{5}=(5/8)\alpha_{4}-25/(8p)
 \quad \mbox{with $p$ in mm.}
 \end{equation}
Moreover, we recall the geometric constraint defined in
Eqs.~(\ref{eq11}) and (\ref{eq21}) which can be rewritten as
\begin{subequations}
\begin{eqnarray}
 g_{1}\!\!\!&=&\!\!\! \frac{1}{\pi}-\eta \leq 0  \\
 g_{2}\!\!\!&=&\!\!\! \alpha_{4}-\frac{1}{2} < 0 \\
 g_{3}\!\!\!&=&\!\!\! \alpha_{4}-\frac{1}{p\kappa_{p \rm max}} < 0 \\
 g_{4}\!\!\!&=&\!\!\! \alpha_{4}-\eta+\frac{b}{p} \leq 0 \\
 g_{5}\!\!\!&=&\!\!\! \alpha_{5}-\frac{1}{4} <0
\end{eqnarray}
\label{eq420}
\end{subequations}
with $\cos^{2}\delta_{i}$ and $\alpha_{5}$ given in
Eqs.~(\ref{eq412}) and (\ref{eq418}), respectively.
\subsection{Results of the Optimization Problem}
We solve the foregoing optimization problem using the $fmincon$ function of Matlab which is base on semi quadratic programming, using $p=50$~mm and $b=9.5$~mm. One
solution is found, corresponding to $\eta=0.69$ and
$a_{4}=24.9992$~mm, with a value of $z=249$. The algorithm finds
the values of $\eta$ and $a_{4}$ as big as possible considering
the constraints. For this solution, constraints (\ref{eq420}d \&
e) are active. Nevertheless, as we saw in Subsection~3.3. about
the influence of parameters $h$ and $a_{4}$, we want these
parameters to be as small as possible in order to have low
pressure angle values. For the solution found above, the service
factor is equal to 0\%. Consequently, we must find a compromise
between the pressure angle and the roller-pin elastic deformation.
Table~\ref{tab401} shows solutions found by the optimization
algorithm upon reducing the boundaries of $\eta$. Each time the
algorithm finds the corresponding value of $a_4$ as big as
possible, constraint (\ref{eq420}d) becomes active. Recorded in
this table is also the corresponding maximum elastic deformation
of the roller pin $v_{L \rm max}$ (its expression is derived
below), the minimum and the maximum absolute values of the
pressure angle, $|\mu_{\rm min}|$ and $|\mu_{\rm max}|$,
respectively, and the service factor, as defined in Section~2.4.
From Eqs.~(\ref{eq400}) and (\ref{eq405}) we have
  \beqa
    v_{L}=\frac{\beta}{a_{5}^{4}}\sqrt{f_{x}^{2}+f_{y}^{2}}
    \nonumber
  \eeqa
Using Eqs.~(\ref{eq404}) and (\ref{eq407}), the above equation
leads to
 \beqa
 v_{L}=\frac{\beta
 F_{0}}{a_{5}^{4}}\sqrt{1+\frac{1}{\tan^{2}\delta}}     \nonumber
 \eeqa
which can be simplified by means of the expression for
$\cos^{2}\delta_{i}$ given in Eq.~(\ref{eq412}) as
 \begin{equation}
 \label{eq423}
 v_{L\rm max}=\frac{\beta F_{0}}{a_{5}^{4}} \frac
{\sqrt{(2\pi \eta-1)^{2}+(\psi_{i}-\pi)^{2}}}{|\psi_{i}-\pi|}
 \end{equation}
with $F_{0}$, $\beta$ and $a_{5}$ given in Eqs.~(\ref{eq404}),
(\ref{eq405}) and (\ref{eq416}), respectively, and
$\psi_{i}=\pi-\Delta$. In Table~\ref{tab401} we record the value
of $v_{L\rm max}$ with $L=10$~mm, $\tau=1.2$~Nm (according to the
Orthoglide specifications recalled in Section~1) and $E=2 \times
10^{5}$~MPa. We conclude from Table~\ref{tab401} that for this cam
profile we cannot find an acceptable compromise between a low
deformation of the roller pin, and hence a high stiffness and
accuracy of the mechanism, and low pressure angle values. Indeed,
for an acceptable deformation of the roller pin, $v_{L {\rm
max}}=8.87~\mu$m, obtained with $\eta=0.38$, the service factor
equals 54.68\%, which is too low. On the other hand, for an
acceptable service factor of 79.43\%, obtained with $\eta=1/\pi
\approx 0.3183$, the deformation of the roller pin is equal to
710.19~$\mu$m, which is too high.

Figure~\ref{fig_two_cams} depicts a rapid protote of the optimum
mechanism whith an intermediate value, namely, $\eta=0.35$. The
motor and the two cams translate, while the follower and the
rollers stay fixed on the base.
\begin{table}[ht]
\begin{center}
{
\begin{tabular}{|c|m{1.7cm}|m{1.7cm}|m{1.5cm}|m{1.8cm}|m{1.6cm}|m{1.6cm}|m{2.7cm}|}
\hline $\eta$  &  $a_{4}$ (mm)  &  $a_{5}$ (mm)  &  $z$ & $v_{L\rm
max}$ ($\mu$m)  &  $|\mu_{\rm min}|$ ($^{\circ}$)  &  $|\mu_{\rm
max}|$ ($^{\circ}$)  &service factor (\%)\\ \hline
 0.69 & 24.99 & 12.50 & 249 & 0.09   & 42.11 & 80.68 & 0 \\
 0.5 & 15.5 & 6.56 & 2968 & 0.50 & 28.59 & 69.81 & 6.85 \\
 0.4 & 10.5 & 3.44 & 32183 & 4.32  & 20.31 & 57.99 & 46.68 \\
 0.39 & 10 & 3.12 & 45490 & 6.07 & 19.46 & 56.42 & 50.68 \\
 0.38 & 9.5 & 2.81 & 66659 & 8.87 & 18.61 & 54.78 & 54.68 \\
 0.37 & 9 & 2.50 & 102171 & 13.63 & 17.75 & 53.04 & 58.69 \\
 0.36 & 8.5 & 2.19 & 165896 & 22.31  & 16.89 & 51.22 & 62.69\\
 0.35 & 8 & 1.87 & 290765 & 39.71 & 16.03 & 49.31 & 66.70\\
 0.34 & 7.5 & 1.56 & 566521 & 79.18 & 15.17 & 47.31 & 70.72 \\
 0.33 & 7 & 1.25 & 1.29 10$^{6}$ & 186.06  & 14.31 & 45.21 & 74.73 \\
 $1/\pi$ & 6.41 & 0.88 & 4.68 10$^{6}$ & 710.19  & 13.31 & 42.64 & 79.43\\
 \hline
\end{tabular}
}
 \end{center}
 \caption{Results of the optimization problem, with
$p=50$~mm, $b=9.5$~mm, $L=10$~mm, $\tau=1.2$~Nm and $E=2 \times
10^{5}$~MPa}
 \label{tab401}
\end{table}
 \begin{figure}[htb]
   \begin{minipage}[b]{8cm}
     \begin{center}
        \psfig{file=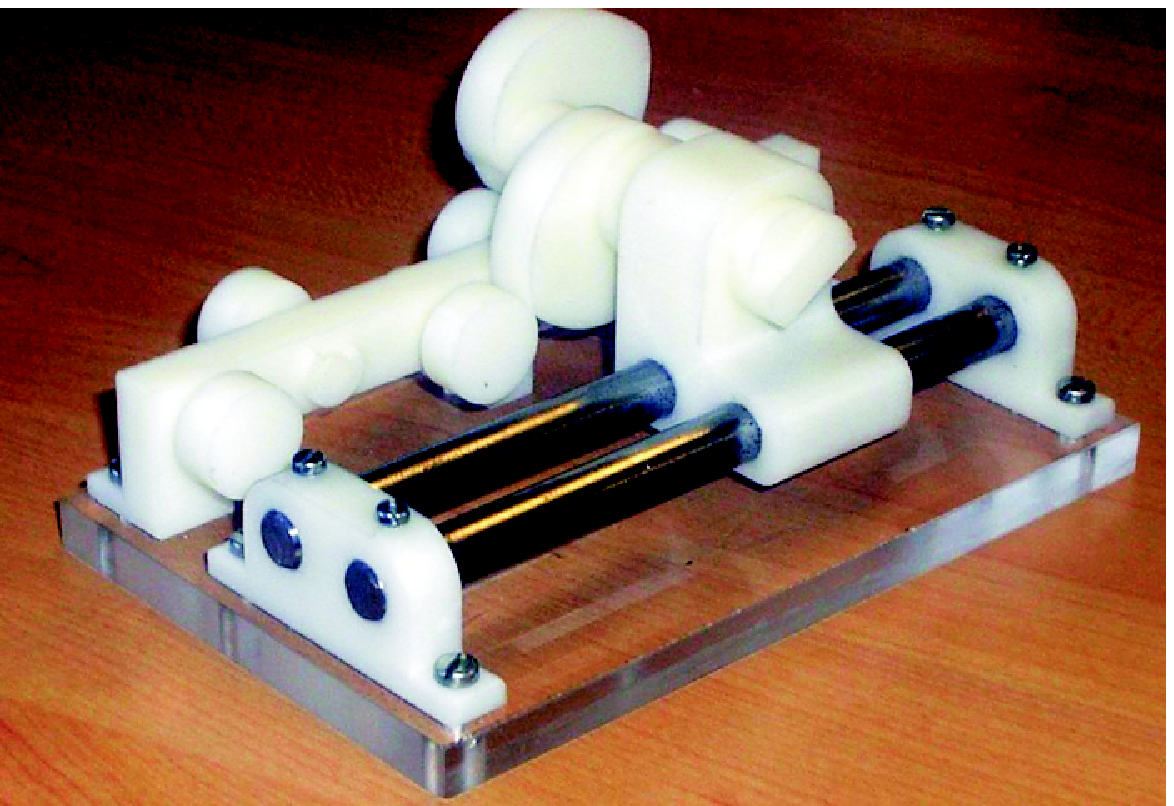, scale=0.5}
        \caption{Layout of the two coaxial conjugate-cam mechanism on the same shaft}
        \label{fig_two_cams}
     \end{center}
   \end{minipage}
   \begin{minipage}[b]{8cm}
   \begin{center}
     \psfig{file=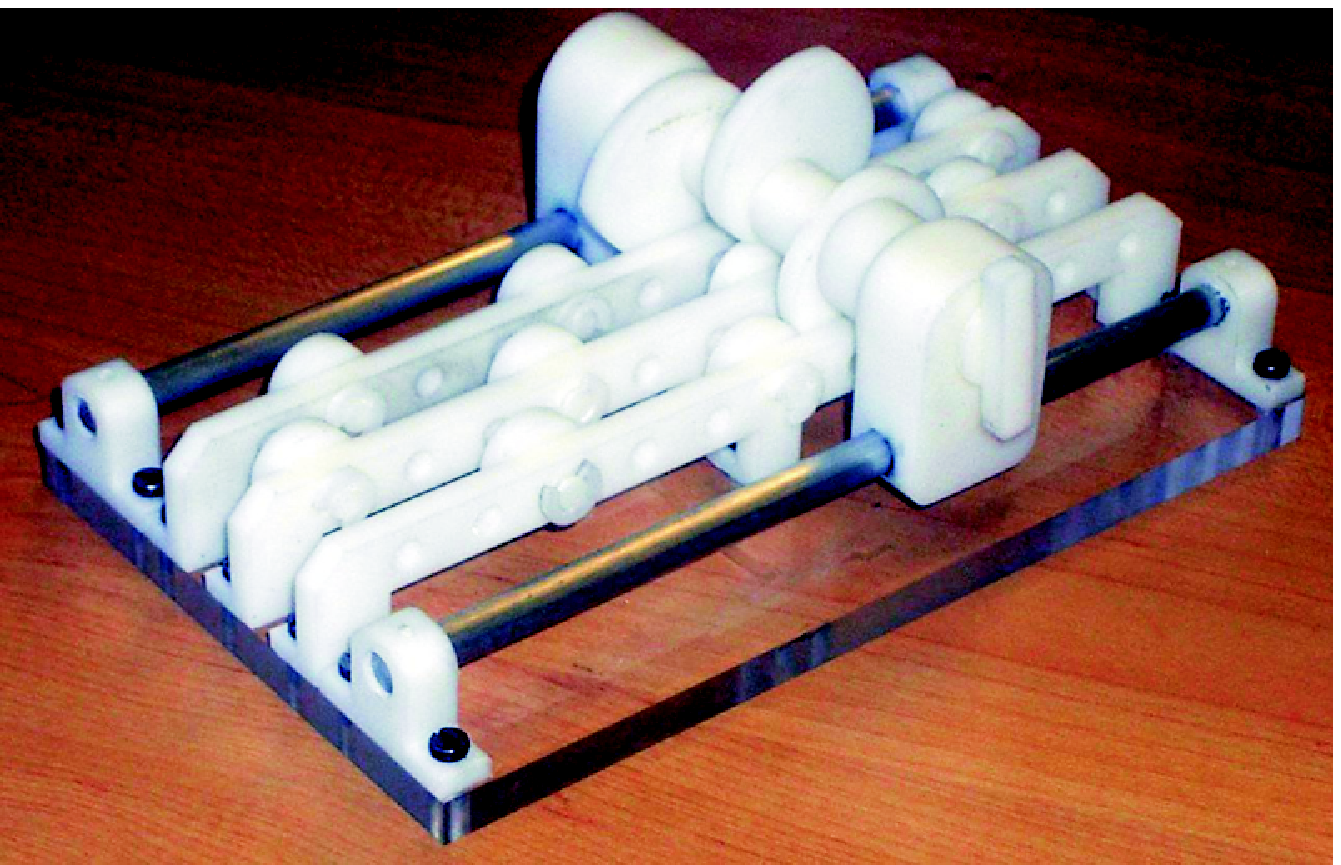, scale=0.46}
     \caption{Layout of the three conjugate-cam mechanism on the same shaft}
     \label{fig_tree_cams}
    \end{center}
   \end{minipage}
 \end{figure}
\section{Novel Conjugate-Cam Mechanism}
This Section describes a novel mechanism and its prototype, based on
Slide-o-Cam, that enables us to decrease considerably the
pressure angle while meeting the Orthoglide specifications.

This mechanism is composed of three conjugate cams mounted either
($i$) on three parallel shafts (with the rollers placed on one
single side of the roller-carrier) or ($ii$) on a single shaft
(with the rollers placed on one single side of three parallel
roller-carrier), as depicted in Fig.~\ref{fig_tree_cams}. Mounted
on two guideways, one motor provides the torque to one camshaft.
In case ($i$), the torque is transmitted to the two other
camshafts through a parallelogram mechanism coupling them, whose
detailed design is reported in \cite{Renotte:2003}. Shown in
Fig.~\ref{fig600} is a layout of this case, with 1, 2 and 3
denoting the three cams.

\begin{figure}[hb]
 \center
 \psfrag{p}{$p$}  \psfrag{p/2}{$p/2$}
 \psfrag{1}{1}  \psfrag{2}{2}  \psfrag{3}{3}
 \psfrag{O1}{$O_1$}
 \psfrag{xu}{$x, u$}  \psfrag{yv}{$y, v$}
 \psfrag{u}{$u$}  \psfrag{v}{$v$}
 \psfrag{s(2Pi/3)}{$s(2 \pi / 3)$}  \psfrag{s(4Pi/3)}{$s(4 \pi / 3)$}
 \psfrag{y12}{$y_{12}$} \psfrag{y13}{$y_{13}$}
 \psfrag{120}{$120^\circ$}
 \psfig{file=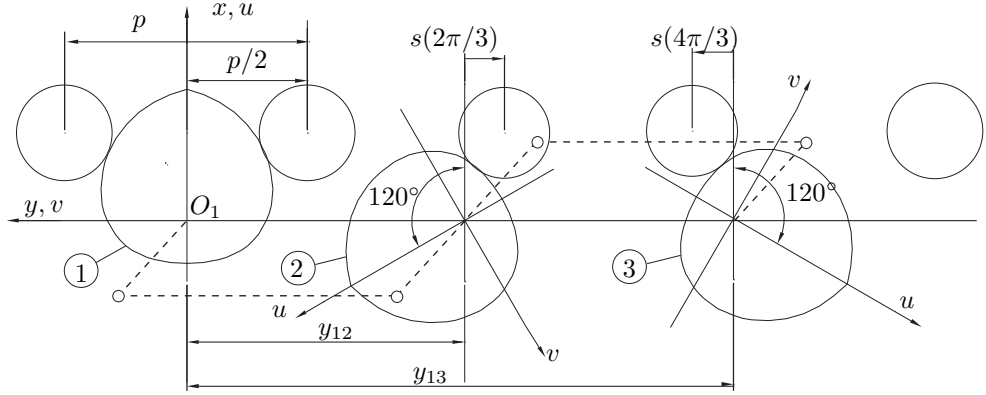, scale=0.6}
 \caption{Layout of the non-coaxial conjugate-cam mechanism on three parallel shafts}
\label{fig600}
\end{figure}
The profile of each cam is as described in Section~2. The cams are
mounted in such a way that the angle between the $u$-axis of cams
1 and 2 is 120$^{\circ}$, while the angle between the $u$-axis of
cams 1 and 3 is 240$^{\circ}$. According to the configuration of
the mechanism depicted in Fig.~\ref{fig600}, and denoting by
$y_{12}$ and $y_{13}$ the distance between the origin of 1 and 2,
and between the origin of 1 and 3, respectively, we have
 \beqa
 y_{12} = p/2+p+s(2\pi/3)~,~~~ y_{13} = p/2 + 2p + s(4\pi/3) \nonumber
 \eeqa
Using the expression of the input-output function $s$ given in
Eq.~(\ref{eq01}), we obtain
 \be
 \label{eq601} y_{12} = 4p/3~,~~~ y_{13} = 8p/3
 \ee
Figure~\ref{fig601} shows the pressure angle variation for each
cam vs.\ $\psi$, where \textit{1}, \textit{2} and \textit{3}
denote the plot of the pressure angle for cams 1, 2 and 3,
respectively. Moreover, cams \textit{2} and \textit{3} are rotated
by angles $2\pi /3$ and $4\pi/3$, respectively, from cam
\textit{1}. We can also consider that the plot for cam 3 that
drives the follower before cam 1 in a previous cycle, referred to
as \textit{3'} in Fig.~\ref{fig601}, is out of phase $-2\pi/3$
with respect to cam \textit{1}. As we saw in Eq.~(\ref{eq01111}),
cam 1 can drive the follower within $\pi \leq \psi \leq 2\pi
-\Delta$, which corresponds in Fig.~\ref{fig601} to the part of
the plot \textit{1} between points $B$ and $D$.

Consequently, cam 2 can drive the follower within
 \begin{eqnarray*}
 \pi+ 2\pi/3 \leq &~\psi~& \leq 2\pi -\Delta+ 2\pi/3
 \quad i.e. \quad
 5\pi / 3 \leq \psi \leq 8\pi/3-\Delta
 \end{eqnarray*}
 {\vspace{-0.1cm}}
and cam 3 within
 {\vspace{-0.1cm}}
 \begin{eqnarray*}
 \pi+ 4\pi/3 \leq &~\psi~& \leq 2\pi -\Delta+ 4\pi/3
 \quad i.e. \quad
 7\pi/3 \leq \psi \leq 10\pi/3-\Delta
 \end{eqnarray*}
which is equivalent to saying that cam 3 can drive the follower in
a previous cycle, within
\begin{eqnarray*}
 \pi-2\pi/3 \leq &~\psi~& \leq 2\pi -\Delta-2\pi/3
 \quad i.e. \quad
 \pi/3 \leq \psi \leq 4\pi/3-\Delta
\end{eqnarray*}

The above interval corresponds in Fig.~\ref{fig601} to the part of
the plot \textit{3'} between points $A$ and $C$. Consequently,
there is a common part for cams 1 (plot \textit{1}) and 3 (plot
\textit{3'}) during which these two cams drive the follower,
namely, between points $B$ and $C$, which corresponds to
 \be
 \pi \leq \psi \leq 4\pi/3-\Delta \label{eq602}
 \ee

Moreover, during this common part, cam 3 has lower absolute
pressure angle values than cam 1, and hence, we consider that only
cam 3 drives the follower. Consequently, cam 1 drives the follower
only within $4\pi/3-\Delta \leq \psi \leq 2\pi-\Delta
\label{eq603}$. These boundaries allow us to have a pressure angle
lower than with coaxial conjugate cams, since we do not use
anymore the part of the cam profile within $\pi-\Delta \leq \psi
\leq 4\pi/3-\Delta$, which has high absolute pressure angle
values. We can thus obtain a higher service factor for the
mechanism.
 \begin{figure}[!hb]
 \begin{center}
 \psfrag{mu}{$\mu$}
 \psfrag{3p}{\textit{3'}} \psfrag{1}{\textit{1}} \psfrag{2p}{\textit{2}} \psfrag{3}{\textit{3}}
 \psfrag{20}{20}   \psfrag{40}{40}   \psfrag{60}{60}   \psfrag{80}{80}   \psfrag{0}{0}
 \psfrag{-20}{-20} \psfrag{-40}{-40} \psfrag{-60}{-60} \psfrag{-80}{-80}
 \psfrag{A}{$A$}  \psfrag{B}{$B$}  \psfrag{C}{$C$}  \psfrag{D}{$D$}
 \psfrag{-2}{-2} \psfrag{-4}{-4}
 \psfrag{2}{2} \psfrag{4}{4}  \psfrag{6}{6}  \psfrag{8}{8} \psfrag{10}{10}
 \psfig{file=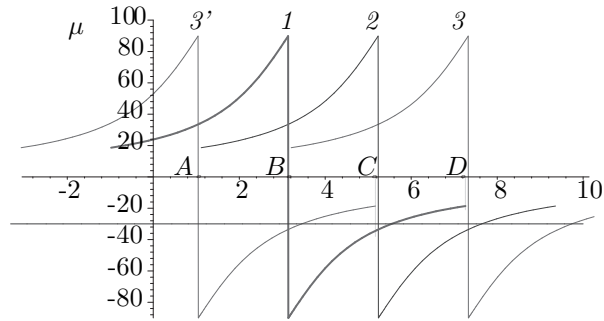, scale=0.5}
 \caption{Pressure angle for the three cams} \label{fig601}
 \end{center}
\end{figure}
\begin{table}[!ht]
\begin{center}
{
\begin{tabular}{|m{0.7cm}|m{1.8cm}|m{1.8cm}|m{1.8cm}|m{1.8cm}|m{1.8cm}|m{3cm}|}
\hline $\eta$  &  $a_{4}$ (mm)  &  $a_{5}$ (mm)  & $v_{L \rm max}$
($\mu$m)  & $|\mu_{\rm min}|$ ($^{\circ}$)  &  $|\mu_{\rm max}|$
($^{\circ}$)  &service factor (\%)\\  \hline
 0.5 & 15.5 & 6.56 & 0.26 & 28.59 & 49.41 & 10.49 \\
 0.4 & 10.5 & 3.44  & 2.88 & 20.31 & 37.20 & 70.02 \\
 0.39 & 10 & 3.12 & 4.14 & 19.46 & 35.81 & 76.02 \\
 0.38 & 9.5 & 2.81 & 6.20 & 18.61 & 34.39 & 82.02 \\
 0.37 & 9 & 2.50 & 9.76 & 17.75 & 32.95 & 88.03 \\
 0.36 & 8.5 & 2.19  & 16.39 & 16.89 & 31.48 & 94.04 \\
 0.35 & 8 & 1.87  & 29.89 & 16.03 & 29.98 & 100 \\
 0.34 & 7.5 & 1.56  & 61.07  & 15.17 & 28.47 & 100 \\
 0.33 & 7 & 1.25  & 147.02 & 14.31 & 26.93 & 100 \\
 $1/\pi$ & 6.41 & 0.88  & 576.95 & 13.31 & 25.12 & 100 \\ \hline
\end{tabular}}
\end{center}
\caption{Design parameters, roller pin deformation and
pressure angle for the novel conjugate-cam mechanism of Fig.~\ref{fig_tree_cams}, with
$p=50$~mm, $b=9.5$~mm, $L=10$~mm, $\tau=1.2$~Nm and $E=2 \times
10^{5}$~MPa.} \label{tab600}
\end{table}

The maximum roller-pin deformation $v_{L \rm max}$ derived in
eq.(\ref{eq423}) is reasonably low. In Table~\ref{tab600}, we
record the values of $\eta$, $a_{4}$, $a_{5}$, $v_{L \rm max}$,
$|\mu_{\rm min}|$, $|\mu_{\rm max}|$ and the service factor for
the three-conjugate-cam mechanism of Fig.~\ref{fig_tree_cams}. The best compromise is to use
the three-conjugate-cam mechanism with $\eta=0.37$, whence the
radius of the roller is $a_{4}=9$~mm and the roller-pin
deformation is $v_{L \rm max}=9.76~\mu$m with a good service
factor of 88.03\%.
\section{Conclusions}
The optimum design of Slide-o-Cam was made for two types of
mechanism, i.e. with two and three-conjugate-cams. The
three-conjugate-cam mechanism reported here allows us to drive the
Orthoglide with prismatic actuators using rotary DC motors.
Moreover, the roller-pin deformation is minimized, and the service
factor is substantially increased with respect to the layout of
Fig.~\ref{fig_two_cams}. Further research is currently underway
to evaluate the influence of the number of lobes on each cam to
increase the service factor using only two cams.
\bibliographystyle{unsrt}

\end{document}